\documentclass[10pt,twocolumn,letterpaper]{article}

\usepackage{cvpr}              

\definecolor{cvprblue}{rgb}{0.21,0.49,0.74}
\usepackage[pagebackref,breaklinks,colorlinks,allcolors=cvprblue]{hyperref}

\newcommand\blfootnote[1]{%
  \begingroup
  \renewcommand\thefootnote{}\footnote{#1}%
  \addtocounter{footnote}{-1}%
  \endgroup
}

\usepackage{graphicx}
\usepackage{array}
\usepackage{soul}
\usepackage{amssymb}
\usepackage{amsmath}

\usepackage[noend]{algpseudocode}
\usepackage[linesnumbered,ruled,vlined,noend]{algorithm2e}
\usepackage{multirow}

\usepackage{lineno}
\usepackage{colortbl}

\def\paperID{17551} 
\def\confName{CVPR}
\def\confYear{2025}

\title{DivPrune: Diversity-based Visual Token Pruning for Large Multimodal Models}

\author{Saeed Ranjbar Alvar, $\text{Gursimran Singh}^{\dagger}$, $\text{Mohammad Akbari}^{\dagger}$, Yong Zhang \\
Huawei Technologies Canada Co., Ltd. \\
{\tt\small \{saeed.ranjbar.alvar1, gursimran.singh1, mohammad.akbari, yong.zhang3\}@huawei.com}
}

\begin{document}
\maketitle

\begin{abstract}
Large Multimodal Models (LMMs) have emerged as powerful models capable of understanding various data modalities, including text, images, and videos. LMMs encode both text and visual data into tokens that are then combined and processed by an integrated Large Language Model (LLM). Including visual tokens substantially increases the total token count, often by thousands. The increased input length for LLM significantly raises the complexity of inference, resulting in high latency in LMMs. To address this issue, token pruning methods, which remove part of the visual tokens, are proposed. The existing token pruning methods either require extensive calibration and fine-tuning or rely on suboptimal importance metrics which results in increased redundancy among the retained tokens. In this paper, we first formulate token pruning as Max-Min Diversity Problem (MMDP) where the goal is to select a subset such that the diversity among the selected {tokens} is maximized. Then, we solve the MMDP to obtain the selected subset and prune the rest. The proposed method, DivPrune, reduces redundancy and achieves the highest diversity of the selected tokens. By ensuring high diversity, the selected tokens better represent the original tokens, enabling effective performance even at high pruning ratios without requiring fine-tuning. Extensive experiments with various LMMs show that DivPrune achieves state-of-the-art accuracy over 16 image- and video-language datasets. Additionally, DivPrune reduces both the end-to-end latency and GPU memory usage for the tested models. The code is available ${\href{https://github.com/vbdi/divprune}{\text{here}}}^{\diamond}$.
\vspace{-5pt}
\end{abstract}
\blfootnote{$\dagger$ Authors have equal contributions.}
\blfootnote{$\diamond$ \href{https://github.com/vbdi/divprune}{https://github.com/vbdi/divprune}}
\vspace{-10pt}
\section{Introduction}
Following the success of Large Language Models (LLMs) in {language} understanding \cite{chiang2023vicuna, touvron2023llama, achiam2023gpt}, Large Multimodal Models (LMMs)~\cite{llava_1.5,llavanext,llavanextvideo,video-llava} have emerged to handle diverse data types like images and video, by leveraging the foundational capabilities of LLMs. 
Typically, LMMs encode text and visual modalities into tokens, also known as embeddings. These tokens are then combined and processed by an integrated LLM. 
{The inclusion of visual tokens significantly increases the total number of tokens, often adding thousands to the combined set. Since the running time and memory requirements scale quadratically with input size ~\cite{transformers_complexity, sukhbaatar2019adaptive, choromanski2020rethinking, dao2022flashattention}, the addition of visual tokens can substantially raise the running time for LMMs}. 
Hence, many of these models often struggle to meet the demands of low-latency applications, particularly in resource-constrained environments \cite{yao2024minicpm}. 

\begin{figure}[tb!]
    \centering
    \includegraphics[width=0.9\linewidth]{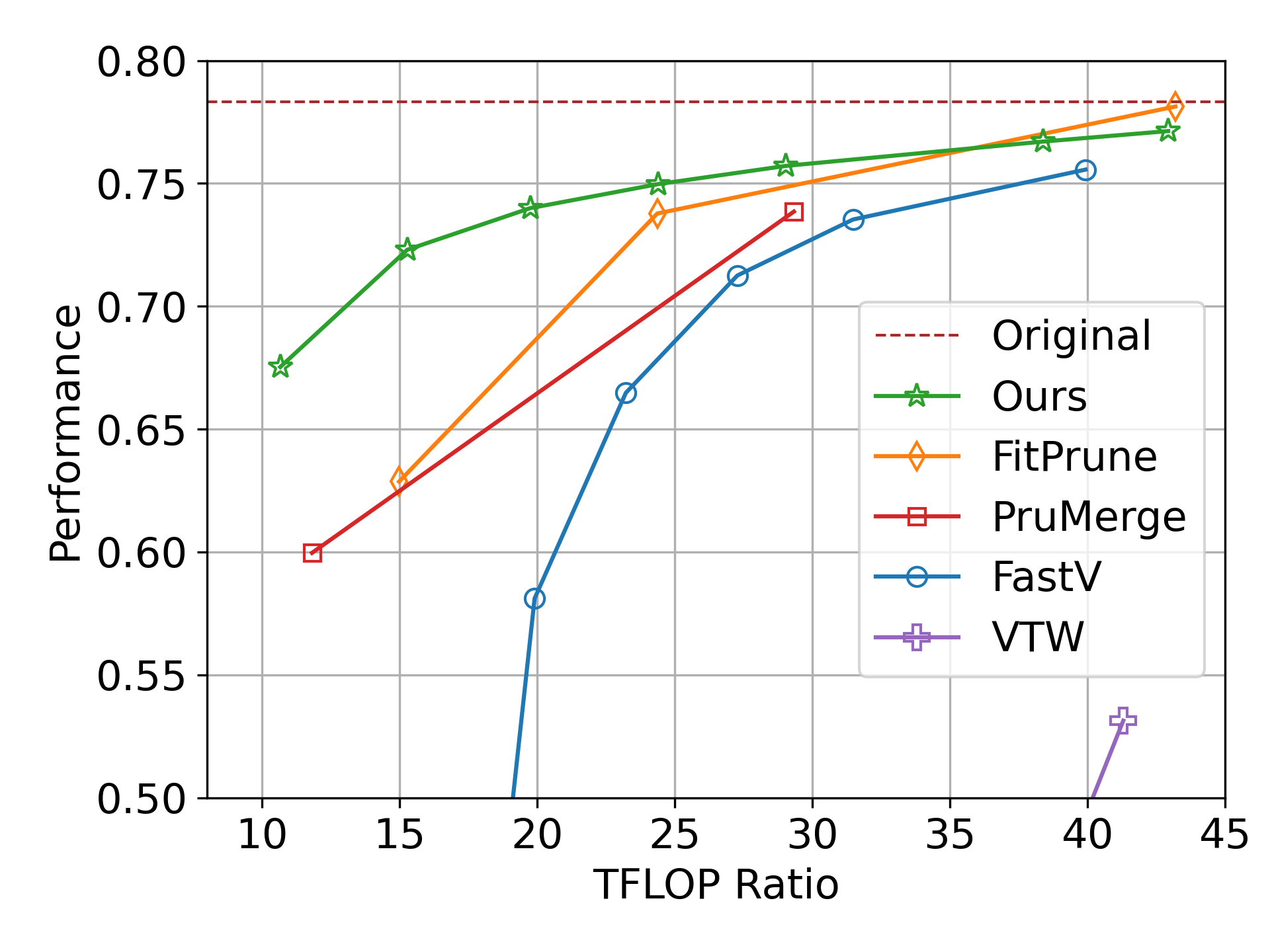}
    \vspace{-5pt}
    \caption{
    {Comparison of different visual token pruning methods across various pruning ratios for LLaVA 1.5-7B. The y-axis is the performance averaged on COCO (CIDEr), OKVQA (Acc), POPE (F1), and MMBench (Acc). The x-axis is the TFLOP ratio of the model after token pruning compared to the original model before pruning. The proposed method significantly outperforms all baselines. Note that, unlike other methods, FitPrune uses an additional calibration step to prune tokens.} 
    \vspace{-10pt}
    }
    \label{fig:tradeoff}
\end{figure}

Previous research~\cite{FastV, FitPrune, PruMerge} has demonstrated that there is a high degree of redundancy in the visual information processed by LMMs. 
As a result, visual token pruning has emerged as a promising solution to address the computational complexity challenges faced by LMMs. Specifically, 
previous research has demonstrated that reducing the number of visual tokens by 50\%~\cite{FastV} to 95\%~\cite{PruMerge} can significantly enhance the inference speed of LMMs.

While promising, token pruning methods have certain shortcomings. {For example, the works in}~\cite{m3, tockenpacker, vtw,FitPrune} require calibration or finetuning for each model which is costly and time-consuming to implement.  
FastV~\cite{FastV} and PruMerge~\cite{PruMerge} use attention scores to identify less important tokens for pruning. However, it is shown that using attention scores is not optimal, as some important tokens are overlooked~\cite{vtw}. 
Additionally, attention-based pruning tends to retain tokens that are similar to each other, leading to redundancy. At high compression ratio, 
this redundancy prevents the selection of a sufficient number of unique tokens to accurately represent the original tokens. 
 In line with this observation, our findings indicate that pruning a large portion of visual tokens using these methods, without subsequent fine-tuning, results in a significant drop in the performance of LMMs across various tasks (Fig.~\ref{fig:tradeoff}).

To address {the above-mentioned} issues, we formulate token pruning as a Max-Min Diversity Problem (MMDP)~\cite{mmdp_def}. In an MMDP, the objective is to select a subset of elements such that the diversity among them is maximized. We apply this concept to token pruning, which we call DivPrune, aiming to maximize the diversity of the selected tokens 
by increasing the minimum distance between them. By ensuring high diversity, DivPrune captures a broader range of visual tokens, making it inherently more robust compared to attention-based methods that focus only on token importance scores. Increasing the diversity also helps ensure that the selected tokens better represent the original {set of} tokens, enabling effective performance even at high pruning ratios without the need for fine-tuning.

DivPrune also offers practical advantages that make it a highly useful solution in real-world scenarios. DivPrune is a plug-and-play solution that can be used without requiring offline optimization with a calibration set, or fine-tuning of the model, which are often time-consuming and computationally expensive. DivPrune is applicable to LMMs with any LLM architecture and vision encoder. Additionally, DivPrune is compatible with inference optimization techniques, such as KV caching, 
resulting in practical speedup in real-world applications. In summary, our major contributions are as follows:

\begin{itemize}
\item We introduce DivPrune, a token pruning method based on MMDP that maximizes diversity among visual tokens, effectively reducing redundancy and ensuring a highly representative subset.
\item DivPrune is a training-free, calibration-data-free, plug-and-play solution that can be seamlessly integrated with off-the-shelf LMMs.
\item We conduct evaluations using 16 datasets on image- and video-language models with image and video understanding tasks. DivPrune achieves state-of-the-art performance, with noticeable gains under extreme pruning (i.e., ratio $\geq$ 80\%).
\item 
DivPrune reduces GPU memory usage and inference latency while maintaining comparable accuracy compared to the original model across most datasets.
\end{itemize}

\section{Related Works}
\subsection{Large Multimodal Models (LMMs)}
{LMMs handle diverse data types, including text, audio, image, and, video~\cite{llava_1.5, llavanext, llavanextvideo, video-llava,  videollama2,gpt40,gemini}}. {This work focuses on open-source LMMs that support language and visual inputs. These LMMs can be categorized into two types: image-based and video-based LMMs. The image-based LMMs~\cite{llava_1.5,llavanext} address image-language understanding tasks, like image captioning, visual question answering, and image reasoning. On the other hand, video-based LMMs are geared towards video understanding~\cite{video-llava,llavanextvideo} tasks, like video captioning, video summarization, and video question answering.}

\subsection{Efficient LMMs}
Several techniques are proposed to improve inference efficiency specifically for LMMs. The first technique is to change the model architecture in LMMs. For example, ~\cite{vl-mamba} proposed to replace transformer-based LLMs with Mamba model~\cite{mamba}.~\cite{tinygpt, tinyllava} retrained LMMs with small scale LLMs 
to improve their efficiency.
~\cite{kd} used knowledge distillation to train a small LMM. In addition to changing the architecture, it is shown in~\cite{skipping} that skipping some blocks or layers within LMMs can improve the inference speed without damaging the model's performance. Furthermore, efficient decoding techniques such as speculative decoding are proposed to make LMM inference more efficient~\cite{sd}.

\subsection{Visual Token Pruning}

Visual token pruning methods are proposed to reduce the inference complexity for LMMs. The first group of methods uses attention scores 
to prune tokens~\cite{PruMerge,FastV}. PruMerge~\cite{PruMerge} introduces a token pruning method for the vision encoder where the visual tokens are clustered and merged based on their attention sparsity. In addition, FastV~\cite{FastV} prunes tokens within a specific layer of the LLM based on the magnitude of attention scores in an earlier layer. 
It is shown that pruning tokens based on attention scores are not optimal~\cite{vtw, attention_is_not}, {especially at higher pruning ratios}.

Calibration-based methods offer another line of work, where pruning layers and/or ratios are determined by analyzing the LLM outputs for a calibration dataset~\cite{FitPrune,vtw}. For example, FitPrune~\cite{FitPrune} calculates a pruning recipe based on the 
observed attention divergence before and after pruning.  VTW~\cite{vtw} argues that visual tokens can be entirely removed after a certain layer within LLM. The layer to remove the visual tokens is chosen using a calibration dataset.
These methods rely on calibration datasets and require custom calibration for each LMM, which can be costly and cumbersome for new models.

Some previous works proposed token pruning with the need for fine-tuning. 
$\text{M}^{3}$~\cite{m3} applies model fine-tuning to produce nested visual token representations at multiple granularities, allowing users to select token lengths dynamically during inference.
In~\cite{tockenpacker}, a projector layer trained using a large-scale dataset is proposed that packs finer detailed information into compact token representations. These methods need significant computational resources for training, limiting their use across various scenarios.

\begin{figure}[tb!]
    \centering
    \includegraphics[width=\linewidth]{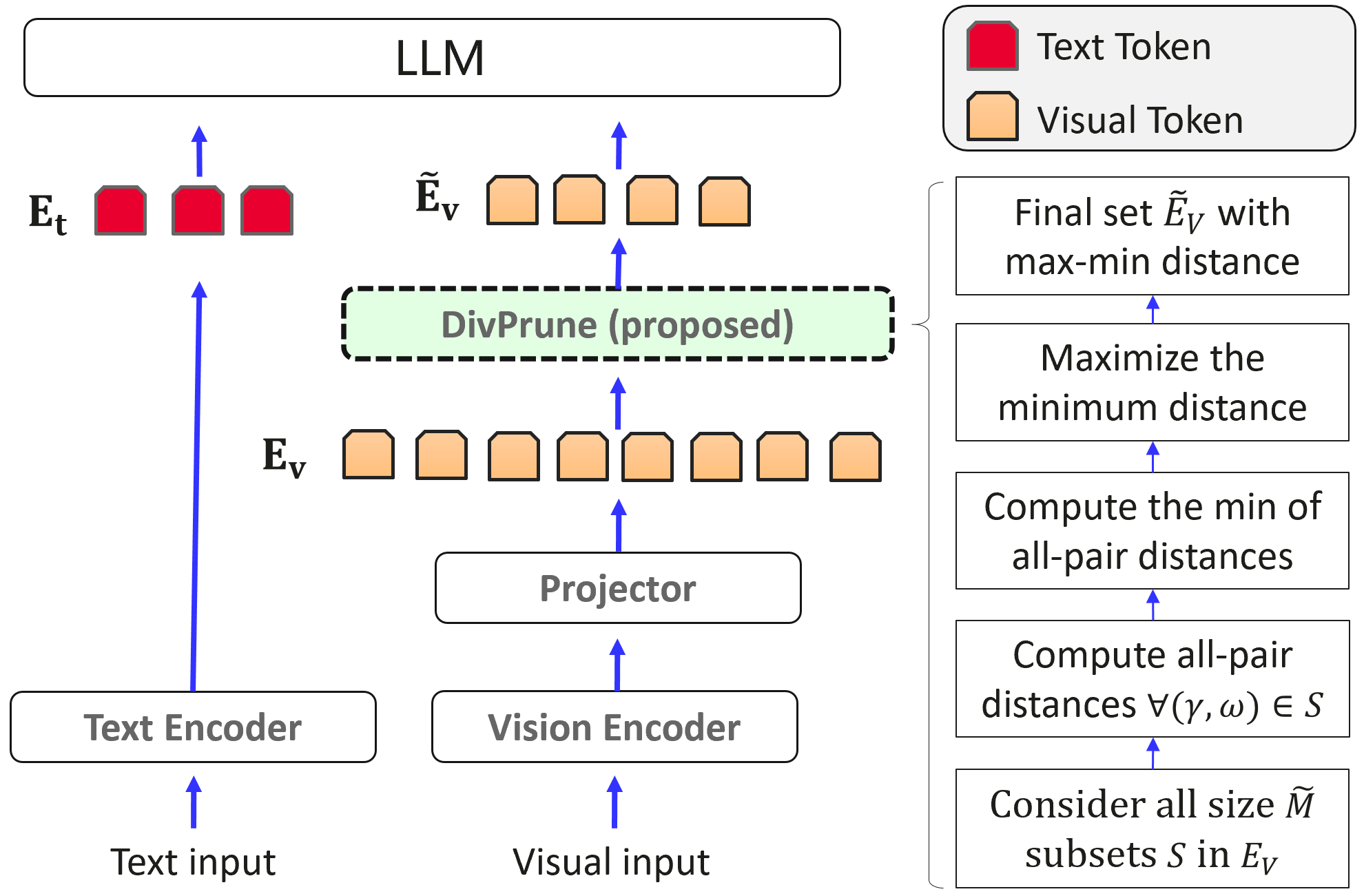}
    \vspace{-10pt}
    \caption{An overview of the LMM architecture, with DivPrune applied to visual tokens. The blocks on the right-hand side illustrate the steps of the method.}
    \vspace{-10pt}
    \label{fig:framework}
\end{figure}

\section{Proposed Method}


In this section, we briefly discuss how LMMs work. Then, the token pruning problem is defined, followed by a detailed presentation of the proposed method.

\subsection{Large Multimodal Models (LMMs)}
An LMM typically processes a pair of inputs, denoted as $(T,V)$, where $T$ is the text input and $V$ is the visual input such as image or video. The text input 
is mapped to $N$ textual {tokens $\mathbf{E_t}=\{t_1,\dots,t_N\}$ using a text encoder.}
{Similarly, the visual input is processed by a corresponding vision encoder. Specifically, it takes visual information $V$ as input and outputs image features, that are further converted to $M$ (generally $M \gg N$) vision tokens $\mathbf{E_v}=\{v_1,\dots,v_M\}$ using a projector layer} (Fig.~\ref{fig:framework}).

The textual {tokens} and visual {tokens} are then  combined to be fed 
to an LLM 
to generate the prediction in an auto-regressive manner. Specifically, $\hat{N}$ output tokens $\mathbf{Y}=\{y_1,\dots,y_{\hat{N}}\}$ are generated as follows:
\begin{equation}
P(y_1, \ldots, y_{\hat{N}} \mid \mathbf{E_t}, \mathbf{E_v}) = \prod_{i=1}^{\hat{N}} P(y_i \mid y_{<i}, \mathbf{E_t}, \mathbf{E_v} ),
\end{equation}
where $P(|)$ is the conditional probability obtained at the output of the LLM.


\subsection{Token Pruning}
Reducing the number of input tokens in an integrated LLM within LMMs helps to lower memory usage and inference latency. 
Since visual {tokens} tend to have more redundancy, they are generally selected for pruning. 

In this context, the problem of token pruning can be defined as follows: given a set of visual tokens \(\mathbf{E_v}\) with $|\mathbf{E_v}|=M$ and the subset size $~\tilde{M}$ ($~\tilde{M}<M$), the goal is to select a subset, $\mathbf{\tilde{E}_v}$, while preserving key information necessary for accurate predictions. To mathematically formulate the token pruning problem, we define a mapping function \(f\), which maps the original set of visual {tokens}, \(\mathbf{E_v}\), to a subset, \(\mathbf{\tilde{E}_v} = \{\tilde{v}_1, \dots, \tilde{v}_{\tilde{M}}\}\), where \(|\mathbf{\tilde{E}_v}| = \tilde{M}\). The objective is to identify a mapping function  $f$ that minimizes the difference in the model's output before and after pruning while ensuring the reduced set still captures the essential information from the original set:
\begin{equation}
\begin{aligned}
\text{Find:} \quad f: \mathbf{E}_v \rightarrow \mathbf{\tilde{E}_v} \\
\text{Objective:} \quad \min_f \, \mathcal{L}\big(\mathcal{P}, \tilde{\mathcal{P}} \big) \\
 \text{Subject to:} \quad |\mathbf{\tilde{E}_v}| = \tilde{M}, \\ 
\end{aligned}
\label{eq:mapping}
\end{equation}
where {$\mathcal{P} = P(y_1, \ldots, y_{\hat{N}} \mid \mathbf{E_t}, {\mathbf{{E}_v}})$ and $\tilde{\mathcal{P}} = P(y_1, \ldots, y_{\hat{N}} \mid \mathbf{E_t}, f(\mathbf{E_v}))$}.
Here, \(\mathcal{L}\) represents a loss function that measures the difference in the model's output with and without pruning, and \(\tilde{M}\) indicates the number of retained {tokens}. 
Next, we propose a novel diversity-based solution for the introduced token pruning problem.



\subsection{DivPrune: Method Overview}
\label{subsec:method}
We proposed a diversity-based token pruning method by re-formulating the problem in \eqref{eq:mapping} {to select} a subset of $\tilde{M}$ elements that maximizes the diversity, thereby reducing redundancy. Specifically, 
we define token pruning as Max–Min Diversity Problem (MMDP)~\cite{mmdp_formula} where the goal is to find the set $\mathbf{\tilde{E}_v}$ among all possible sets  with $\tilde{M}$ samples in $\mathbf{{E}_v}$ that has the maximum minimum distance between 
its elements. So, MMDP is defined as: 
\begin{equation}
\text{Find } \mathbf{\tilde{E}_v}{=} {\text{ arg max}} \left[  \min_{\gamma,\omega\in S}{\big(d(\gamma,\omega)\big): \forall S\subset \mathbf{{E}_v}}\right],
\label{eq:mmdp}
\end{equation}
where $S$ is an arbitrary set in $\mathbf{
{E}_v}$ with $\tilde{M}$ elements and $(\gamma,\omega)$ are arbitrary elements in $S$. The distance is measured by $d(.,.)$ which is defined using the cosine distance as follows:
\begin{equation}
{ d(\gamma, \omega) } = 1 - \frac{\mathbf{\gamma} \cdot \mathbf{\omega}}{\|\mathbf{\gamma}\| \|\mathbf{\omega}\|}.
\end{equation}

A solution for the MMDP problem in \eqref{eq:mmdp} is a subset of $\mathbf{E_v}$ that maximizes diversity by minimizing redundancy between elements. In the literature, several solutions including exact and heuristic methods are proposed to solve the MMDP problem~\cite{mmdp_def, mmdp_solutions}. Since the number of {tokens} is generally limited (e.g., 576 in LLaVA 1.5~\cite{llava_1.5}) and the solvers are not generally designed for GPU acceleration, we obtain exact solution for the problem. Notably, the overhead of the selection process using GPU is negligible compared to the computations within the LLM. Detailed steps of the proposed method is summarized in Algorithm~\ref{alg:bruteforce}. Once the selected {tokens} are identified, the remaining visual {tokens} are discarded. The selected {tokens} along with the textual {tokens} are passed to the LLM. 
 
As shown in \cref{{alg:bruteforce}}, the proposed method has two stages after the initialization. The selected subset, $\mathbf{\tilde{E}_v}$, is initialized as empty, and the candidate list $\mathbf{R}$ is initialized with all the visual tokens. In the first stage, the first token of the selected subset is chosen based on the pairwise distance between the tokens of the candidate list. 
Then, the chosen token is moved from the candidate list to the selected list. 
In the second stage, similar to the first stage, the pairwise distance of the tokens in $\mathbf{\tilde{E}_v}$ and the tokens in $\mathbf{R}$ is used to add samples to $\mathbf{\tilde{E}_v}$ iteratively. Finally, once the number of tokens in  $\mathbf{\tilde{E}_v}$ reaches the specified subset size, the selection procedure is terminated and the  $\mathbf{\tilde{E}_v}$ is returned. 
To avoid repeated distance calculations over iterations a distance matrix is initially calculated by one matrix multiplication. 

\def\NoNumber#1{{\def\alglinenumber##1{}\State #1}\addtocounter{ALG@line}{-1}}

\SetKwRepeat{Do}{do}{while}%
\begin{algorithm}[tb!]
\small
\caption{Proposed Token Pruning Method}
\label{alg:bruteforce}
\SetAlgoLined
 $\tilde{M}$: subset size; $\mathbf{{E}_v}$: visual tokens; $\mathbf{\tilde{E}_v}$: selected subset
 
 Initialize $\mathbf{\tilde{E}_v}$=[] and  $\mathbf{R} = \mathbf{{E}_v}$ 
 
\textcolor{gray}{// First stage: add the first token}

    $D = []$ initialize the distance array 

    \For{$i$ {in}  $\mathbf{R}$}{  
    
    $d_{min} = +inf$
    
    \For{$j$ {in}  $\mathbf{R}$}{
        \hspace{3pt} \textbf{If} {$\big(i\neq j {~\&~} d(i,j) \leq d_{min}\big)$ \textbf{then} $d_{min} = d(i,j)$}
     }
    Add $d_{min}$ to $D$ 
    }

    $k = \mathbf{R}[\text{arg max (D)}]$   
    
    move  $k$ from $\mathbf{R}$ to $\mathbf{\tilde{E}_v}$
 
 \textcolor{gray}{// Second stage: iteratively add the subsequent tokens}
 
 \While{$|\mathbf{\tilde{E}_v}| < \tilde{M}$}{
 
    $D = []$ initialize the distance array  

    \For{$i$ {in}  $\mathbf{R}$}{  
    
    $d_{min} = +inf$
    
    \For{$j$ {in}  $\mathbf{\tilde{E}_v}$}{
        \hspace{3pt} \textbf{If} {$d(i,j) \leq d_{min}$ \textbf{then} $d_{min} = d(i,j)$}
     }
    Add $d_{min}$ to $D$ 
    }

    $k = \mathbf{R}[\text{arg max (D)}]$   
    
    move  $k$ from $\mathbf{R}$   to $\mathbf{\tilde{E}_v}$  
}
Return $\mathbf{\tilde{E}_v}$
\end{algorithm}
The proposed method can also be applied to the features (i.e., hidden states) in the intermediate layers of the LLM. In this case, our method is not applied to the visual {tokens}, but to the features corresponding to the visual {tokens} obtained from a decoder layer to select a subset before feeding them to the subsequent layers. In either case, 
our method obtains the highest diversity for the selected elements. Ablation studies are provided in the next section to analyze the effect of pruning different elements at different layers. 


\section{Experiments}
In this section, we present a comprehensive analysis {comparing the performance of} our method {and previous works} across various settings, tasks, and datasets. Insights into the proposed method are also provided through illustrative examples. {Moreover, the efficiency of DivPrune along with ablation study are provided.}

\subsection{Experimental Settings}
\textbf{Baselines and Models}: We consider five baselines, namely, FastV~\cite{FastV}, PruMerge~\cite{PruMerge}, VTW~\cite{vtw}, FitPrune~\cite{FitPrune} and $\text{M}^3$~\cite{m3}. Among these, we consider FastV, PruMerge, and VTW as our {main competitors} as they are plug-and-play and do not rely on any further costly finetuning or calibration process. However, for the sake of completeness, we also report performance comparison with respect to one finetuning-based ($\text{M}^3$) and one calibration-based (FitPrune) methods. 
{Note that, VTW, by default, requires calibration to determine the best layer for a given task. However, doing that does not allow us to set a specific TFLOP ratio, complicating the comparison. Hence, whenever required we disable the calibration of VTW to select the layer that matches the FLOP requirement of a particular experiment. }

We test DivPrune and the baselines with popular LMMs namely LLaVA 1.5-7B~\cite{llava_1.5}\footnote{https://huggingface.co/liuhaotian/llava-v1.5-7B}, LLaVA 1.5-13B~\cite{llava_1.5}\footnote{https://huggingface.co/liuhaotian/llava-v1.5-13b} 
LLaVA 1.6-7B\footnote{https://huggingface.co/liuhaotian/llava-v1.6-vicuna-7b} (also known as LLaVA-NeXT~\cite{llavanext}), 
and LLaVA-NeXT-Video-7B~\cite{llavanextvideo}\footnote{https://huggingface.co/lmms-lab/LLaVA-NeXT-Video-7B-DPO} to demonstrate the generality of DivPrune. 
{For each tested model and task, we report only the relevant subset of baseline that is applicable to that specific model and task, alongside our results.} 

{All the tested LMMs used CLIP vision encoder~\cite{clip}. LLaVA 1.5 model uses 576 visual tokens to represent images. LLaVA 1.6 converts each image into a varying number of patches, resulting in  3-5 times more visual tokens compared to LLaVA 1.5.  LLaVA-NeXT-Video uses 144 tokens to process each frame. For all the experiments with LLaVA-NeXT-Video we used a total of 8 frames resulting in 1152 tokens for the processed frames.} 
 
\textbf{Datasets, Tasks, and Metrics}: We selected a comprehensive set of common tasks and datasets aimed at multimodal reasoning and understanding. Specifically, we chose 11 image-language and 5 video-language datasets. These datasets encompass a wide range of tasks, including captioning, multiple-choice Question Answering (QA), and open-ended QA based on text and image/video inputs. Consistent with prior works, CIDEr score~\cite{cider} is used for evaluating captioning tasks, and Exact Match (EM), Accuracy (Acc), F1, Perception Score (P-score)~\cite{mme} and GPT-assisted~\cite{gptscore} score are used for QA tasks. Furthermore, Wu-Palmer similarity (WUPS) score~\cite{wups} and GPT-assisted  score~\cite{gptscore} is used for open-ended QA. For all task performance metrics used in this paper, higher values indicate better performance. For the reported time and memory, lower values indicate better results. Further details regarding the datasets, tasks, and metrics are provided in the supplementary material. 


Following the earlier works in ~\cite{FastV, FitPrune, vtw}, we report the computational requirement, measured in TFLOPs, for DivPrune and the baselines. Various configurations including different pruning ratios at different layers are examined to obtain different working TFLOPs for our method and the baselines. The reported TFLOP ratio is the TFLOP of the model with pruned tokens relative to the original model's TFLOP with no pruning. This ratio is estimated as~\cite{FastV}: 
\begin{equation}
\resizebox{.9\hsize}{!}{ 
$\frac{K \times (4 \mu d^2 - 2 \mu^2 d + 2\mu dm) + (T-K) \times (4 \tilde{\mu} d^2 - 2 \tilde{\mu}^2 d + 2 \tilde{\mu} dm)}{T \times (4 \mu d^2 - 2 \mu^2 d + 2 \mu dm)}$,
}
\end{equation}

where $T$ is the total transformer-based decoder layers. $\mu = N + M$ is the total sequence length before pruning, $\tilde{\mu} = N + \tilde{M}$ is the sequence length after pruning, $d$ is the hidden state size of the layer, and ${m}$ is the intermediate size of feed-forward network module. 
{Depending on the TFLOP ratio requirement set by a particular experiment, we adjust the pruning hyperparameters of all baselines to match that requirement}. However, some baselines do not support fine-grained adjustments like our approach does. In these cases, we choose the smallest available TFLOP ratio that exceeds the requirement set by an experiment, which might give these baselines a slight advantage over our method. 

We used 8 $\times$ V100 GPUs with 32GB VRAM for all the experiments in this paper. Additionally, 
we used the lmms-evals
package~\cite{lmmseval} for running these benchmarks for all the baselines and models. All results are obtained with a batch size of 1. 
For the metrics that require ChatGPT API access, the model is set to ``gpt-4o-mini".


\begin{figure}[bh!]
    \centering
    \begin{subfigure}{\linewidth}
        \centering
        \includegraphics[width=0.95\textwidth]{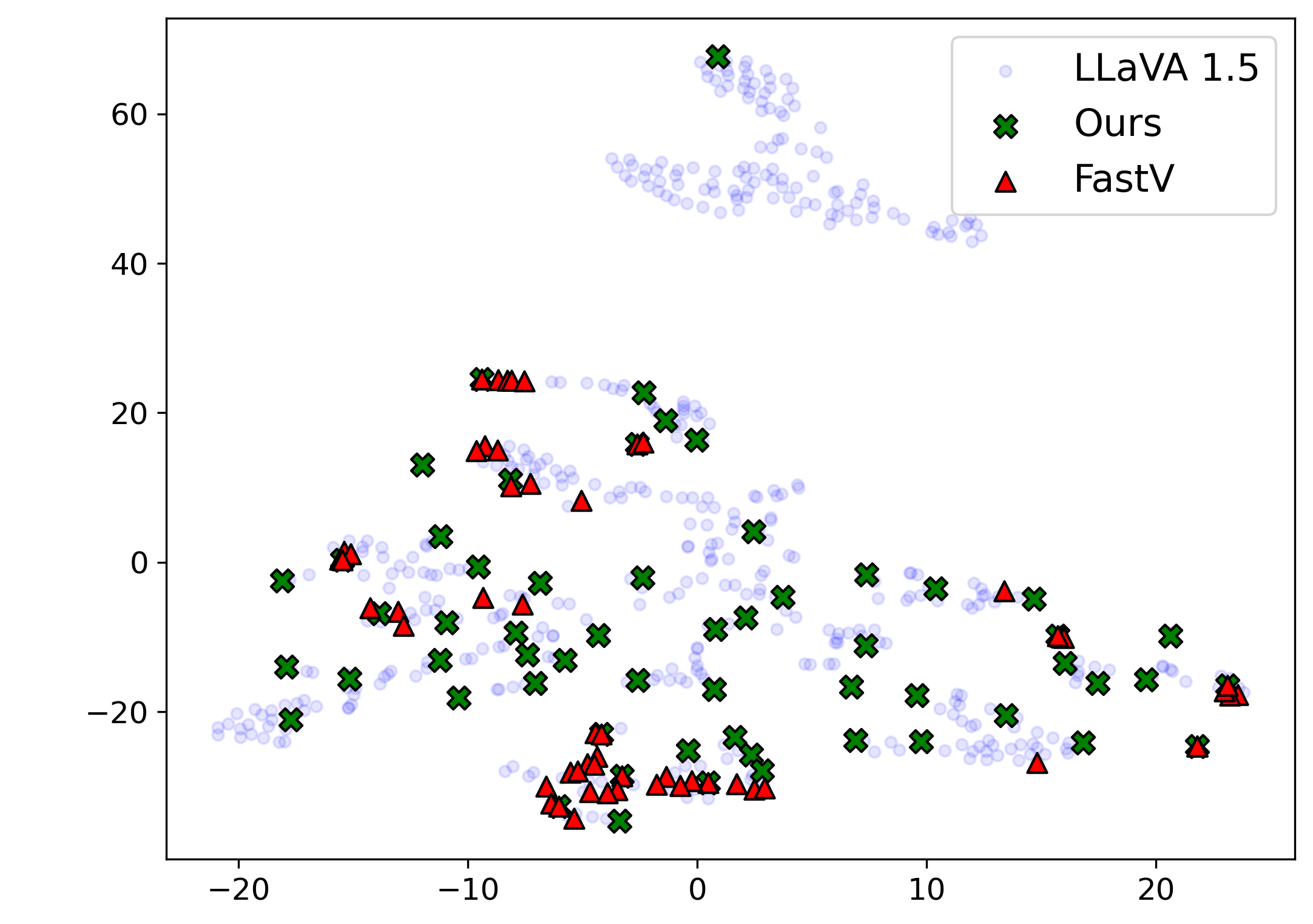}
        \caption{}
        \label{fig:figure1}
    \end{subfigure}
    
    \begin{subfigure}{\linewidth}
        \centering
        \includegraphics[width=0.95\textwidth]{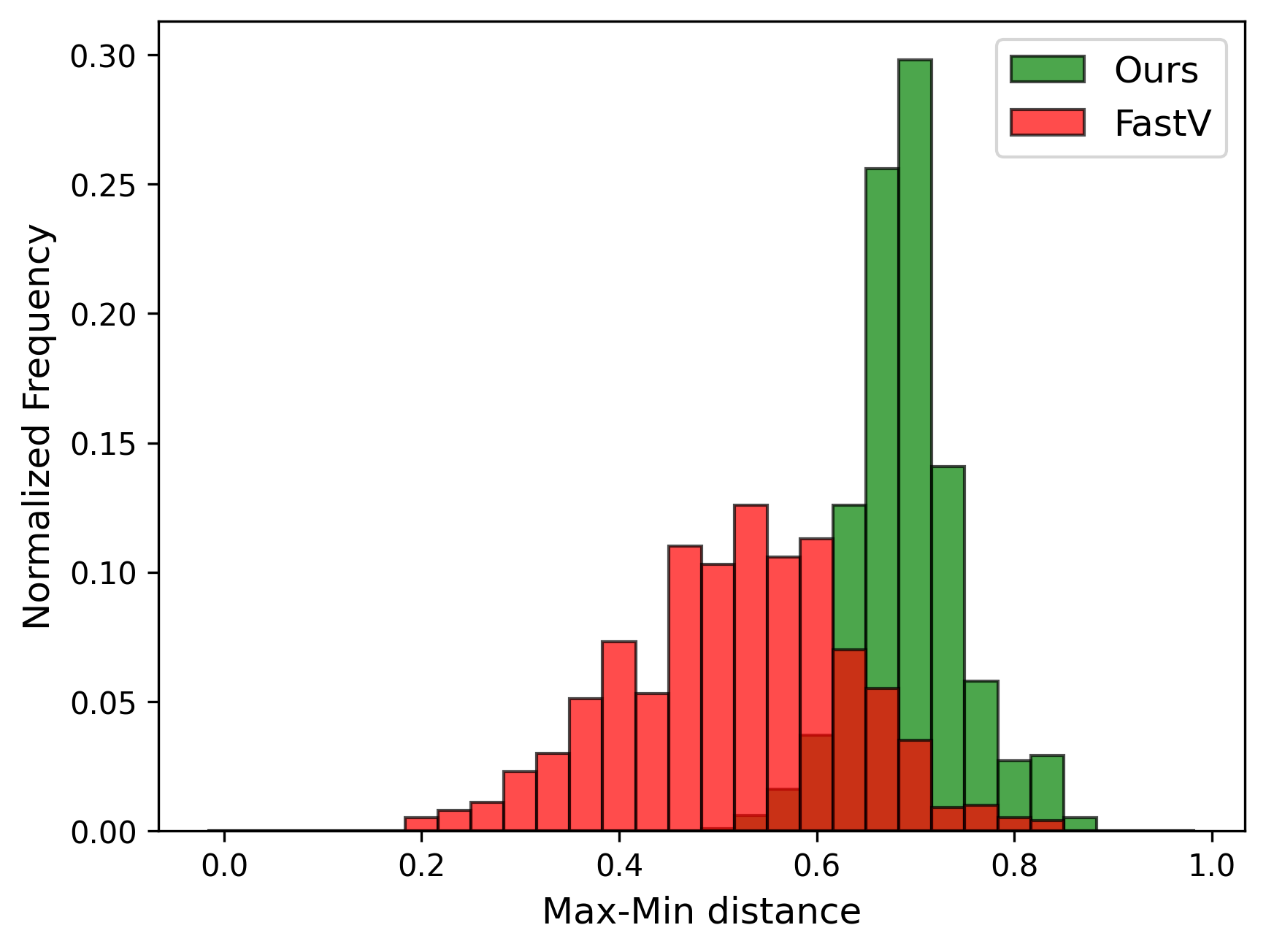}        
        \caption{}
        \label{fig:figure2}
    \end{subfigure}
    \vspace{-15pt}
    \caption{(a) t-SNE visualization of {visual tokens for the original model, our method, and FastV.} (b) Histogram of the Max-Min distance between the selected tokens over the SeedBench dataset. 
    }
    \label{fig:visualization}
\end{figure}

\subsection{Insights}

We provide visualizations comparing DivPrune with importance-based token pruning methods using LLaVA 1.5-7B and the SeedBench dataset~\cite{SeedBench}. Detailed analysis across different models and datasets is provided in the following subsections.

The visual tokens in LLaVa 1.5 model are 4096-dimensional vectors. 
The t-SNE method~\cite{t-sne} is utilized to {project the visual tokens in $\mathbf{{E}_v}$ from a high dimensional to a} 2D space. 
The corresponding visualization for a sample input data is shown in Fig.~\ref{fig:visualization}-(a) using light Pruple points. Then, DivPrune is applied to select 10\% of the visual tokens (i.e., pruning 90\%). Additionally, FastV, as an importance-based token pruning method, which utilizes attention scores, is employed to prune with the same ratio. The selected subsets using DivPrune and FastV are shown with different markers in Fig.~\ref{fig:visualization}-(a). More examples are provided in the supplementary materials. 

\renewcommand{\arraystretch}{1.03} 

\setlength\heavyrulewidth{0.25ex}
\begin{table*}[tb!]
\centering
\small
\resizebox{\textwidth}{!}{%
\begin{tabular}{c|cc|cccccccccccc}
\toprule
\multirow{2}{*}{} & \multirow{2}{*}{Method} & {TFLOP} & COCO  & Flickr & GQA & MMB & MME & MMMU & Nocaps & OKVQA & POPE & SQA  & SEEDB \\ 
    &          &   (ratio~\%)   & CIDEr & CIDEr  & EM    & Acc & P-score & Acc   & CIDEr  & EM    & F1 & EM    & Acc  \\ \midrule \multirow{9}{*}{\rotatebox[origin=c]{90}{\textcolor{black}{\textbf{LLaVA 1.5-7B}}}} 
&\cellcolor[gray]{0.85} Original  & \cellcolor[gray]{0.85} 3.228 (100.00) & \cellcolor[gray]{0.85}1.10  & \cellcolor[gray]{0.85}0.75   & \cellcolor[gray]{0.85}61.96 & \cellcolor[gray]{0.85}64.09       & \cellcolor[gray]{0.85}1506  & \cellcolor[gray]{0.85}36.44 & \cellcolor[gray]{0.85}1.06   & \cellcolor[gray]{0.85}53.39 & \cellcolor[gray]{0.85}85.84     & \cellcolor[gray]{0.85}69.41 & \cellcolor[gray]{0.85}66.17  \\
&VTW~\cite{vtw}& 0.603 (18.46) & 0.05  & 0.03   & 38.94 & 21.31 & 681 & 32.60 & 0.03  & 18.64 & 25.35 & 65.29 & 36.13  \\
&FastV~\cite{FastV} & 0.514 (15.69) & 0.06  & 0.03   & 38.73 & 20.62 & 696   & 32.00 & 0.04 & 18.32 & 32.84   & 65.15 & 35.69  \\
&\textbf{Ours}  & 0.512 (15.63)  & \textbf{0.96}  & \textbf{0.62}   & \textbf{56.85} & \textbf{59.19} & \textbf{1328}  & \textbf{35.89} & \textbf{0.92}   & \textbf{46.98} & \textbf{86.02} & \textbf{68.27} & \textbf{59.47} \\ \cline{2-14}

&PruMerge~\cite{PruMerge} & Variable & 0.77 & 0.50 & 51.30 & 54.47 & 1259 & 35.11 & 0.73   & 41.74 & 66.89  & \textbf{68.91} & 53.26  \\
&$\text{\textbf{Ours}}^{*}$  & Variable & \textbf{0.91}  & \textbf{0.56}  & \textbf{55.25} & \textbf{58.16} & \textbf{1330}  & \textbf{35.44} & \textbf{0.87}   & \textbf{44.38} & \textbf{83.06}  & 67.87 & \textbf{57.88}  \\ \cline{2-14}
&$\text{FitPrune}^{\bigtriangleup}$~\cite{FitPrune}    & 0.513 (15.65)                           & 0.90  & 0.56   & 52.39 & 57.65       & 1197  & 36.00 & 0.86   & 42.53 & 60.89     & 68.02 & 54.84  \\
&$\text{M}^{3\bullet}$~\cite{m3}  & 0.512 (15.63) & 1.00  & 0.67   & 60.81 & 65.81 & 1391  & 31.80 & 0.95 & 55.12 & 86.33 & 64.65 & 64.93  \\
&$\text{PruMerge-LoRA}^{\bullet}$ & Variable & 0.96  & 0.63 & 55.96 & 59.88 & 1334  & 34.89 & 0.90   & 47.99 & 77.13  & 68.32 & 57.93 \\  \midrule \multirow{6}{*}{\rotatebox[origin=c]{90}{\textcolor{black}{\textbf{LLaVA~1.5-13B}}}} 
&\cellcolor[gray]{0.85}Original  &   \cellcolor[gray]{0.85} 6.281 (100.00)  &  \cellcolor[gray]{0.85}1.16 & \cellcolor[gray]{0.85}0.80 & \cellcolor[gray]{0.85}63.33 & \cellcolor[gray]{0.85}68.64 & \cellcolor[gray]{0.85}1522 & \cellcolor[gray]{0.85}35.67 & \cellcolor[gray]{0.85}1.09 & \cellcolor[gray]{0.85}58.28 & \cellcolor[gray]{0.85}85.99 & \cellcolor[gray]{0.85}72.88 & \cellcolor[gray]{0.85}66.82 \\
&VTW~\cite{vtw}                & 1.030 (16.16) & 0.08 & 0.05 & 39.71 & 21.91 & 622 & 32.10 & 0.05 & 22.49 & 0.40 & 66.24 & 38.59 \\
&FastV~\cite{FastV}  & 1.003 (15.73) & 0.38 & 0.18 & 44.98 & 37.80 & 942 & \textbf{35.11} & 0.33 & 32.14 & 30.02 & 69.96 & 44.95 \\
&\text{\textbf{Ours}} & 1.002 (15.71) & \textbf{1.00} & \textbf{0.66} & \textbf{57.29} & \textbf{63.40} & \textbf{1407} & 34.89 & \textbf{0.95} & \textbf{53.29} & \textbf{83.43} & \textbf{72.34} & \textbf{62.04} \\ \cline{2-14}
&$\text{PruMerge}^{\bigtriangleup}$~\cite{PruMerge}  & Variable & 0.80 & 0.53 & 52.01 & 58.93 & 1256 & \textbf{36.56} & 0.77 & 49.15 & 64.36 & \textbf{72.53} & 56.10 \\
&$\text{\textbf{Ours}}^{*}$   & Variable & \textbf{0.94} & \textbf{0.59} & \textbf{56.09} & \textbf{61.77} & \textbf{1344} & 34.89 & \textbf{0.91} & \textbf{50.86} & \textbf{79.60} & 71.34 & \textbf{60.00}\\  \midrule  \multirow{4}{*}{\rotatebox[origin=l]{90}{\textcolor{black}{{\textbf{LLaVA~1.6-7B}}}}}
&\cellcolor[gray]{0.85}Original  &   \cellcolor[gray]{0.85} 11.849 (100.00)  & \cellcolor[gray]{0.85}1.00 & \cellcolor[gray]{0.85}0.68 & \cellcolor[gray]{0.85}64.28 & \cellcolor[gray]{0.85}67.01 & \cellcolor[gray]{0.85}1520 & \cellcolor[gray]{0.85}36.44 & \cellcolor[gray]{0.85}0.88 & \cellcolor[gray]{0.85}44.20 & \cellcolor[gray]{0.85}86.38 & \cellcolor[gray]{0.85}70.15 & \cellcolor[gray]{0.85}70.16 \\  
&VTW~\cite{vtw}  & 1.318 (11.23) & 0.06 & 0.03 & 38.62 & 19.76 & 606 & 31.30 & 0.03 & 8.66 & 7.13 & 65.74 & 37.48 \\
&FastV~\cite{FastV}               & 1.327 (11.30) & 0.06 & 0.03 & 38.79 & 20.36 & 619 & 32.56 & 0.04 & 8.80 & 7.78 & 65.49 & 37.62 \\

&\text{\textbf{Ours}} & 1.266 (10.79) & \textbf{0.89} & \textbf{0.61} & \textbf{58.69} & \textbf{63.49} & \textbf{1362} & \textbf{37.11} & \textbf{0.76} & \textbf{41.92} & \textbf{82.97} & \textbf{68.57} & \textbf{64.11} \\\cline{2-14}
&$\text{M}^{3\bullet}$~\cite{m3}  & 1.266 (10.79) & 1.01 & 0.67 & 62.97 & 69.16 & 1490 & 35.00 & 0.85 & 57.49 & 87.44 & 69.51 & 68.49 \\
\bottomrule
\end{tabular}
}
\vspace{-5pt}
\caption{Comparison results of our method and different baselines on image-language understanding datasets. ${\bullet}$: Finetuning is used, ${\bigtriangleup}$: Calibration dataset is used.  $\text{\textbf{Ours}}^{*}$: Our method matching the PruMerge selection ratio. 
}
\vspace{-5pt}
\label{tbl:benchmark_comparison_image}
\end{table*}

As the example in Fig.~\ref{fig:visualization}-(a) shows, the proposed method selects points from all the clusters that appeared in the projected space whereas FastV does not choose any samples from the upper cluster. So, our method achieves a better representation of the original points by including samples from all clusters. In addition, the FastV method selects many tokens that are very close to each other which increases redundancy among the selected set. On the other hand, our method reduces redundancy by pruning the closely similar tokens.

In addition, the max-min distance (Eq.~\ref{eq:mmdp}) for the selected subset of tokens is computed using 1000 randomly data samples from the SeedBench dataset and the histogram of the computed values is shown in Fig.~\ref{fig:visualization}-(b). As the plot indicates, the proposed method selects a subset where samples have a higher minimum pair-wise distance compared to the FastV method. Hence, our method achieves higher diversity among the selected tokens that have less redundancy compared to the ones chosen using FastV. We analyze the effect of the reduced diversity on task performance in the following sections.

\subsection{Image-Language Understanding}
In this section, we compare DivPrune against baselines across various image-language understanding tasks, including open- and closed-ended QA, visual reasoning, and image captioning. Specifically, ScienceQA-IMG (SQA)~\cite{sqa-i}, POPE~\cite{pope}, MME~\cite{mme}, MMB~\cite{mmb},  GQA~\cite{gqa}, MMMU~\cite{MMMU}, Flicker30k~\cite{flicker}, SeedBench (SEEDB)~\cite{SeedBench}, Nocaps~\cite{nocaps}, OKVQA~\cite{okvqa}, and COCO-2017~\cite{coco} are used. 

In the first experiment, summarized in \cref{tbl:benchmark_comparison_image}, we analyze an extreme compression scenario for three  image-based LMMs by fixing the TFLOP ratio at approximately 15\%, wherever the baseline allows configuration to a fixed TFLOP ratio.
Since PruMerge does not allow fixing the TFLOP ratio, we configure our approach (Ours*) to match the variable pruning corresponding to PruMerge for a fair comparison. 
In the top section of the table, we compare the results of various baselines for LLaVa 1.5-7B. Specifically, the baselines supporting LLaVA 1.5 are grouped into three categories: plug-and-play methods, those with a variable TFLOP ratio, and those requiring a calibration dataset or involving fine-tuning the LMMs. 
Among the plug-and-play methods, which are the focus of this work, our approach significantly outperforms both the VTW and FastV baselines across all datasets. This result holds despite using lower TFLOPs, clearly demonstrating the advantage of our method in this scenario. 
{For instance, when DivPrune is used, the performance of LLaVA 1.5-7b decreases by 5.1\% on the GQA dataset and 4.9\% on the MMB dataset. In contrast, the VTW and FastV methods result in performance drops of at least 23.0\% and 42.8\% on these datasets, respectively. The performance gap between DivPrune and the baseline methods is even more pronounced in image captioning tasks. For example, the CIDEr score on the COCO dataset drops by approximately 95\% with VTW and FastV, but only by 12.7\% with DivPrune. Additionally, DivPrune, compared to the original model, shows less than a 2\% performance drop on the MMMU and SQA datasets and slightly enhances the original model’s performance on the POPE dataset while reducing the TLOP ratio by 84.4\%. It is shown that removing redundant tokens in some datasets can improve the original model's performance~\cite{FastV}.}

\renewcommand{\arraystretch}{1.03} 

\begin{table*}[tb!]
\centering
\resizebox{.89\linewidth}{!}{%
\begin{tabular}{cc|ccccc||ccc}
\toprule
\multirow{2}{*}{}& TFLOPs & ActivityNet & SeedBench & VChatGPT  & NextQA & EgoSch. & Max GPU & Prefill Time & E2E Latency \\ 
                & (ratio \%) & Score/Acc             & Acc            & Score         & WUPS           & Acc       &  mem (GB)           & (sec) & (sec)    \\ \midrule \rowcolor{gray!20}
Original  &    6.539 (100)    & 2.67 / 48.10          & 38.7           & 2.16          & 26.05          & 41.8        & 14.06  & 0.330 & 4.37     \\
VTW~\cite{vtw}      &   1.124 (16.97)     & 1.61 / 26.84          & 29.39          & 1.19          & 18.66          & 25.42    &  13.63 & \textbf{0.150 } & 3.43    \\
FastV~\cite{FastV}      &  0.943 (14.20)    & 1.95 / 33.91 & 32.98  & 1.44  & 22.51          & 29.14 & 13.57 & \textbf{0.150} & 3.63     \\
\textbf{Ours}        &  0.937 (14.10)    & \textbf{2.56 / 45.90} & \textbf{37.00} & \textbf{1.92} & \textbf{24.48} & \textbf{39.76}  &  \textbf{13.51}  & 0.161 & \textbf{3.39}  \\\bottomrule
\end{tabular}
}
\vspace{-5pt}
\caption{Comparison results of our method and baselines on LLaVA-NeXT-Video-7B across video-language understanding datasets. 
}
\vspace{-5pt}
\label{tbl:benchmark_comparison_video}
\end{table*}

Next, in the variable scenario, the pruning ratio is determined dynamically. To ensure a fair comparison, we matched the pruning ratio with that of the PruMerge baseline, assuming the average sequence length for calculating the average TFLOPs across each dataset. As indicated by the results, our approach consistently outperforms PruMerge across all benchmarks, except one. Further, for the baseline with calibration, we observe that our approach outperforms the FitPrune approach on nearly all datasets {by} {up to 25.1\%}, despite not using any calibration dataset.  Finally, compared to baselines involving fine-tuning, our method achieves comparable or superior performance without requiring any fine-tuning.

The above experiment is repeated with LLaVa 1.5-13B model and the results are shown in the middle part of~\cref{tbl:benchmark_comparison_image}. The baselines that support this model are FastV, VTW, and PruMerge. As shown in the table, DivPrune outperforms the corresponding baselines in both plug-and-play and variable scenarios almost on all the tested datasets.  
{For example, on the POPE dataset, DivPrune outperforms VTW, FastV, and PruMerge with F1 score improvements of 83\%, 53.4\%, and 15.2\%, respectively. Additionally, on the MMB dataset, DivPrune achieves higher accuracy rates of 41.5\%, 25.6\%, and 2.8\% compared to VTW, FastV, and PruMerge, respectively. }
This demonstrates that DivPrune generalizes effectively across models with varying numbers of parameters.

{In the bottom part of~\cref{tbl:benchmark_comparison_image}, the results corresponding to LLaVA 1.6-7B model are shown. We used the same pruning ratio as for LLava 1.5. However,  the lower TFLOP ratio is due to the large number of visual tokens in LLaVA 1.6. The results indicate that the performance of the model drops significantly when baseline pruning methods are applied. For example, the F1 score on the POPE dataset drops by 79\% with the baselines as compared to the original model, whereas the drop with DivPrune is only 3.4\%. DivPrune also maintains competitive performance compared to the original model across various datasets. Specifically, DivPrune shows only 3.5\%, 2.3\%,3.4\%,1.6\% drop in accuracy compared to the original model on the MMB, OKVQA, POPE, and SQA datasets, respectively, while reducing the TFLOP by 89\%. The results also demonstrate that pruning visual tokens with DivPrune enhances the original model’s performance on the MMMU task. These results show that DivPrune generalizes across different models. Qualitative examples as well as results with additional datasets are provided in the supplementary materials.}

Furthermore, we show the comparison of different baselines and our method across various TFLOP ratios. We plot the results in Fig.~\ref{fig:tradeoff} where the y-axis represents average performance on four datasets, namely, COCO (CIDEr), OKVQA (Acc), POPE (F1), and MMBench (Acc). The range of the performance metric for all datasets is between 0 and 1, except for the CIDEr metric, which has a maximum reported value of 1.10. On the x-axis, we only show the high compression scenario (TFLOP ratio $\le$ 45\%). As shown in the figure, our method significantly outperforms all the baselines, particularly in high compression scenarios (TFLOP $\le$ 25\%). Further, we notice a steep drop in performance of all baselines as the TFLOP ratio $\rightarrow 10$, while our method falls more gracefully.  This results in an increasing performance gap between our approach and the baselines at extreme compression levels. For higher TFLOP ratios almost all converge toward the original performance, with FitPrune slightly outperforming our approach by an insignificant margin. It is important to note that, unlike our method, FitPrune relies on a calibration dataset to prune tokens.

\subsection{Video-Language Understanding}
In this section, LLaVA-NeXT-Video-7B~\cite{llavanext}, a video-based LMM is used to analyze the performance of the proposed method on various video-language understanding tasks. Specifically, we evaluate DivPrune using five datasets, namely, ActivityNet~\cite{activitynet}, SeedBench~\cite{SeedBench}, VideoChatGPT (temporal) ~\cite{videochatgpt}, NextQA~\cite{nextqa}, and EgoSchema~\cite{egoschema}. FastV and VTW methods are chosen as the baselines. We tested DivPrune using the same pruning ratio as in the image understanding experiments. However, due to the higher number of visual tokens in the LLaVA-NeXT-Video model, this pruning ratio results in lower TFLOPs ratio. For the baselines, we match their TFLOPs with ours by selecting the smallest available TFLOP ratio that exceeds the TFLOPs of our method. The results for the original model, DivPrune, and the baselines are given in \cref{tbl:benchmark_comparison_video}. 
As shown in the table, DivPrune outperforms both FastV and VTW by a significant margin. Specifically, 
DivPrune achieves {upto} 12\% higher accuracy than FastV and {upto} 19\% better than VTW on Video QA datasets including ActivityNet, SeedBench, and EgoSchema. DivPrune also outperforms both baselines on open-ended QA such as VideoChatGPT and NextQA by achieving higher GPT-assisted and WUPS scores.

Furthermore, our method achieves performance that is highly competitive compared to the original model without pruning despite using only 14.1\% of the original model's TFLOPs. This demonstrates the robustness of DivPrune, as it effectively generalizes to video LMMs. Notably, the performance gap between DivPrune and the original model without pruning narrows as the number of visual tokens increases, indicating that DivPrune is more effective for the models with larger visual contexts.

\subsection{Efficiency Analysis}
In this section, we analyze the efficiency of the proposed method using memory usage (i.e., max allocated memory), prefill time, and end-to-end latency (E2E). For this experiment, VideoChatGPT dataset with 499 samples is used to obtain the average time and memory usage for LLaVA-NeXT-Video-7B model. The results are summarized on the right side of \cref{tbl:benchmark_comparison_video}.
The obtained results are compared against the original model, as well as the FastV and VTW baselines. As shown in the table, our approach requires approximately 400MB less memory than the original model, with memory usage comparable to the baselines. In terms of prefill and E2E time, our approach is about 55\% and 22\% faster, respectively, compared to the original model. When compared to the baselines, our prefill time is approximately 6-7\% longer, while the E2E time is 1-7\% shorter. The slight increase in prefill time for our method compared to the baselines is due to the distance calculations (See Section \ref{subsec:method}), which are performed only once during the prefill stage. In contrast, for baselines, the corresponding calculations for token pruning need to be done at each decoding step, resulting in longer E2E time.

\subsection{Ablation Study}
In this section, we conduct an ablation study to analyze the impact of modifying various core components of our method. The ablation experiments are conducted with the LLaVA 1.5-7B model. First, we show the effect of pruning tokens inside the LLM in \cref{tbl:ablations-layers} using 5 datasets. By default, in our method, visual tokens are pruned before being passed to the first decoder layer in the LLM, which we refer to as 'Layer 0'. We also tested 'Layer 1' where the first layer is processed without pruning and the pruning is performed afterward. We further extended this approach by allowing tokens to pass through the first few layers unpruned and then pruning them after specific layers. As shown in the table, for a fixed TFLOP ratio of $19.61\%$, pruning done by our method at layer 0 achieves higher task accuracies compared to pruning at layers 1, 2, and 3 of the LLM.

\renewcommand{\arraystretch}{1}
\begin{table}[tb!]
\centering
\resizebox{\linewidth}{!}{%
\begin{tabular}{cc|ccccc}
\toprule
                                                 & TFLOP & MMB     & MMMU                   & POPE           & SQA & \multirow{2}{*}{Avg}            \\
\multirow{-2}{*}{}  &        (ratio \%)        & Acc    & Acc                     & F1      & EM   &           \\ \midrule
Layer 0 (\textbf{Ours})                                   & {{19.61}}          & \textbf{59.19} & \textbf{35.89}  & \textbf{86.02} & \textbf{68.27} & \textbf{62.34} \\
Layer 1                                     & {19.65}          & 59.02          & 34.89                  & 80.67          & 67.18        & {60.44} \\
Layer 2                                   & {19.70}          & 54.90          & 34.22                   & 69.27          & 69.56         & {56.99}\\
Layer 3                                     & {19.80}          & 23.97          & 32.67                  & 31.82          & 65.94 & {38.60} \\ \bottomrule

\end{tabular}
}
\vspace{-5pt}
\caption{Ablation study on applying DivPrune at different layers.}
\vspace{-5pt}
\label{tbl:ablations-layers}
\end{table}

\begin{table}[tb!]
\centering
\resizebox{\linewidth}{!}{%
\begin{tabular}{cc|cccccc}
\toprule
                             & TFLOP & MMB     & MMMU           & POPE       & SQA & \multirow{2}{*}{Avg}            \\
\multirow{-2}{*}{}  &        (ratio \%)        & Acc    & Acc            & F1     & EM &             \\ \midrule
Cosine (\textbf{Ours}) & 19.61 & 59.19 & \textbf{35.89}  & \textbf{86.02} & \textbf{68.27} & \textbf{62.34}  \\ \midrule
$\ell_1$ &     {19.61}  & 59.71 & 34.67  & 85.40 & 67.97 & 61.94 \\
$\ell_2$ &     {19.61}  & \textbf{59.97} & 35.00  & 85.64 & \textbf{68.27} & 62.22 \\
\midrule
Random &       {19.61}  & 52.66 & 34.56  & 72.78 & 66.63 & 56.66 \\
Min-Max &      {19.61}  & 38.57 & 33.11  & 49.26 & 65.20 & 46.53 \\
\bottomrule
\end{tabular}
}
\vspace{-5pt}
\caption{Ablation on using various diversity measures. 
}
\vspace{-5pt}
\label{tbl:ablations-distmetric}
\end{table}

Furthermore, in \cref{tbl:ablations-distmetric}, we provide an analysis of using alternative diversity measures for token pruning. The first three rows show the impact of choosing different distance measures to quantify the similarity among tokens. It can be seen that all three similarity measures, cosine, $\ell_1$ , and $\ell_2$ perform comparably, with cosine (default setting) performing slightly better. This suggests that the choice of similarity measure does not significantly impact DivPrune's overall performance.

The last two rows in \cref{tbl:ablations-distmetric} show the effect of choosing alternative strategies of token selection other than the proposed \textit{Max-Min} diversity-based solution \eqref{eq:mmdp}. We tested random pruning as well as the Min-Max strategy where the maximum distance between the selected samples is minimized. The Min-Max strategy enforces high redundancy among the selected samples, resulting in reduced diversity. As results in the bottom part of \cref{tbl:ablations-distmetric} reveal that any deviation from our proposed selection strategy results in suboptimal performance. Specifically, the Min-Max strategy performs the worst, showing approximately 15.8\% lower performance compared to ours. This decline is due to the Min-Max approach selecting tokens that are highly similar to each other, resulting in less diversity among the selected visual tokens. Random selection provides some degree of diversity, but it performs 5.6\% worse than the proposed method because it cannot guarantee maximum diversity. This proves that redundancy of visual tokens leads to poor performance and diversity maximization is needed for optimal performance, corroborating the utility and need of the proposed diversity maximization in \cref{eq:mmdp}.

\section{Conclusion}
In this paper, we proposed a token pruning method based on a max-min diversity problem, called DivPrune. In the proposed method, maximum diversity is achieved among the selected tokens, resulting in reduced redundancy. By ensuring high diversity, the selected tokens provide a more representative subset of the original tokens, enabling effective performance even at high pruning ratios without requiring fine-tuning. Extensive experiments were conducted with multiple LMMs on image and video understanding tasks across 16 datasets. The results show that DivPrune achieves state-of-the-art accuracy on the tested datasets. DivPrune generalizes well to different model sizes and architectures, while also improving memory consumption and end-to-end latency for the tested LMMs.




\appendix
\section*{Supplementary Material}

\section{Datasets, Tasks, and Metrics}
We briefly introduce the 11 image-language and 5 video-language datasets used in the experiments  of the main manuscript. 
In addition, the system prompt (instruction) used to get output results for each dataset is given. The details of datasets used for image-language and video-language understanding tasks are presented in Tab.~\ref{tbl:dataset_detail}. 
Furthermore, the details on 3 extra datasets used for our new experiments in the supplementary material are provided.

As shown in the table, diverse range of tasks including image captioning, visual reasoning, open-ended visual question answering, closed-ended visual question answering, and multiple-choice visual question answering are used to evaluate the performance of the visual token pruning methods compared with ours. Note that the system prompts are the default prompts provided in the lmms-evals evaluation package \cite{lmmseval}.

\section{More Examples for Insights}
In Fig.~3 of the main manuscript, DivPrune and an importance-based token pruning method (i.e., FastV~\cite{FastV}) are compared using (a) t-SNE visualization for a sample input's visual tokens and (b) a histogram of the max-min distance between the selected tokens across 1000 data samples from SeedBench dataset \cite{SeedBench}. In this section, additional examples from SeedBench and GQA datasets \cite{gqa} are respectively provided in Fig.~\ref{fig:visualization_supp}-(a)-(b) and Fig.~\ref{fig:visualization_supp}-(c)-(f).

As shown in Fig.~\ref{fig:visualization_supp}-(a)-(b), similar to the observation in the main manuscript, the majority of the selected tokens using FastV method are densely clustered near each other, whereas the tokens selected using DivPrune are more widely separated. As a result, the redundancy among the selected tokens decreases. In addition, unlike DivPrune, FastV does not include any tokens from the top clusters. Hence, DivPrune achieve a better representation for the original set of tokens.

Further examples using GQA dataset are provided in Fig.~\ref{fig:visualization_supp}-(a)-(e). Inline 
with earlier observation, Divprune reduces redundancy and achieves better representation compared to importance-based token pruning when applied to GQA dataset. To verify this behavior over multiple dataset samples, the max-min distance among the selected visual tokens is obtained using 1000 randomly selected samples from the GQA. The histogram of the obtained max-min values for DivPrune and FastV is shown Fig.~\ref{fig:visualization_supp}-(f). The histogram also verifies that our method achieves higher max-min distance values, thereby reducing redundancy for the tested samples of the dataset.

\renewcommand{\arraystretch}{1.03} 
\setlength\heavyrulewidth{0.25ex}
\begin{table}[b!]
\centering
\resizebox{\linewidth}{!}{%
\begin{tabular}{c|cc|ccc}
\toprule
\multirow{2}{*}{} & \multirow{2}{*}{Method} & {TFLOP}  & TextVQA & VizWiz & VQAv2\\ 
    &          &   (ratio~\%)   & EM & EM & EM  \\ \midrule \multirow{7}{*}{\rotatebox[origin=c]{90}{\textcolor{black}{\textbf{LLaVA 1.5-7B}}}} 
&\cellcolor{gray!20} Original & \cellcolor{gray!20} 3.13 (100.00) & \cellcolor{gray!20} 46.08  & \cellcolor{gray!20} 54.24  & \cellcolor{gray!20} 76.65  \\
&VTW~\cite{vtw}& 0.507 (16.20) & 8.22 & 50.13 &   42.13\\
&FastV~\cite{FastV} & 0.418 (13.35) & 8.21 & 50.48 & 41.71\\
&\textbf{Ours} & 0.416 (13.29) & \textbf{35.97} & \textbf{57.41} & \textbf{71.55} \\ \cline{2-6}

&PruMerge~\cite{PruMerge} & Variable &  \textbf{37.70} & 56.31 & 65.01  \\ 
&$\text{\textbf{Ours}}^{*}$  & Variable & 35.00 & \textbf{57.43} & \textbf{69.59}  \\ \cline{2-6}
&$\text{FitPrune}^{\bigtriangleup}$~\cite{FitPrune} & 0.417 (13.32) & 30.10 & 54.62 & 64.86 \\ 
&$\text{M}^{3\bullet}$~\cite{m3} & 0.416 (13.29) & 44.31 & 52.98  &  75.87  \\\bottomrule
\end{tabular}
}
\caption{Comparison results of our method and baselines on three additional datasets. ${\bullet}$: Finetuning is used, ${\bigtriangleup}$: Calibration dataset is used.  $\text{\textbf{Ours}}^{*}$: Our method matching the PruMerge selection ratio.}

\label{tbl:additional_dataset}
\end{table}

\renewcommand{\arraystretch}{1.05} 

\setlength\heavyrulewidth{0.25ex}
\begin{table*}[tbh!]
\small
\resizebox{\textwidth}{!}{%
\begin{tabular}{c|cccl}
\toprule
& Dataset               & Task              & Metric & System Prompt \\ \midrule \multirow{13}{*}{\rotatebox[origin=l]{90}{\textcolor{black}{\textbf{Image-Language Understanding}}}} 
&COCO-2017~\cite{coco}             & Image Captioning  & CIDEr                                                                 & Provide a one-sentence caption for the provided image.                                                                                                                     \\ \cline{2-5}
&Flicker30k~\cite{flicker}           & Image Captioning  & CIDEr                                                                 & Provide a one-sentence caption for the provided image.                                                                                                                     \\\cline{2-5}
&GQA~\cite{gqa}                   & CE-VQA            & Eaxct Match                                                           & Answer the question using a single word or phrase.                                                                                                                         \\\cline{2-5}
&MMBench~\cite{mmb}               & MC-VQA            & Accuracy                                                              & Answer with the option's letter from the given choices   directly.                                                                                                         \\\cline{2-5}
&MME~\cite{mme}                   & CE-VQA            & Perception Score                                                      & Answer the question using a single word or phrase.                                                                                                                         \\\cline{2-5}
&MMU~\cite{MMMU}                   & CE-VQA and OE-VQA & Accuracy                                                             & \begin{tabular}[c]{@{}l@{}}Answer with the option's letter from the given choices   directly, OR \\ Answer the question using a single word or phrase.\end{tabular}        \\\cline{2-5}
&Nocaps~\cite{nocaps}                & Image Captioning  & CIDEr                                                                 & Provide a one-sentence caption for the provided image                                                                                                                      \\\cline{2-5}
&OKVQA~\cite{okvqa}                & Visual Reasoning  & Exact Match                                                           & \begin{tabular}[c]{@{}l@{}}When the provided information is insufficient, respond with   'Unanswerable'.\\ Answer the question using a single word or phrase.\end{tabular} \\\cline{2-5}
&POPE~\cite{pope}                  & CE-VQA            & F1 Score                                                              & Answer the question using a single word or phrase.                                                                                                                         \\\cline{2-5}
&ScienceQA-Image~\cite{sqa-i}       & Visual reasoning  & Exact Match                                                          & Answer with the option's letter from the given choices   directly.                                                                                                         \\\cline{2-5}
&SeedBench-Image~\cite{SeedBench}        & MC-VQA            & Accuracy                                                              & Answer with the option's letter from the given choices   directly.                                                                                                         \\ \cline{2-5}

&TextVQA~\cite{textvqa}   &    CE-VQA         &   Exact Match         & Answer the question using a single word or phrase. \\ \cline{2-5}
&VizWiz~\cite{vizwiz}    &      CE-VQA       &     Exact Match       & \begin{tabular}[c]{@{}l@{}}When the provided information is insufficient, respond with   'Unanswerable'.\\ Answer the question using a single word or phrase.\end{tabular}  \\ \cline{2-5}
&VQAv2~\cite{vqa_v2}     &    CE-VQA         &    Exact Match       & Answer the question using a single word or phrase. \\ 
\midrule \midrule \multirow{4}{*}{\rotatebox[origin=c]{90}{\textcolor{black}{\textbf{~~Video-Language}}}} 
&ActivityNet~\cite{activitynet}           & CE-VQA            & \begin{tabular}[c]{@{}c@{}}Accuracy/\\ GPT-Assisted score\end{tabular} & Answer the question using a single word or phrase.                                                                                                                         \\\cline{2-5}
&SeedBench-Video~\cite{SeedBench}       & MC-VQA            & Accuracy                                                              & Answer with the option's letter from the given choices   directly.                                                                                                         \\\cline{2-5}
&VideoChatGPT-temporal~\cite{videochatgpt} & OE-VQA            & GPT-Assisted-score                                                    & Evaluate the temporal accuracy of the prediction compared to   the answer.$^*$                                                                                                 \\\cline{2-5}
&NextQA~\cite{nextqa}                & CE-VQA            & WUPS                                                                  & Answer a question using a short phrase or sentence.                                                                                                                        \\\cline{2-5}
&EgoSchema~\cite{egoschema}            & MC-VQA            & Accuracy                                                              & Answer with the option's letter from the given choices   directly.   \\     
\bottomrule
\end{tabular}
}
\caption{Details of the datasets, the corresponding tasks, metrics, and prompts used in our experiments. CE-VQA: Closed-Ended Visual Question Answering, OE-VQA: Open-Ended Visual Question Answering, MC-VQA: Multiple-Choice Visual Question Answering. $^*$: Only the main sentence from the prompt is shown here.}
\label{tbl:dataset_detail}
\end{table*}

\begin{figure}[t!]
    \centering
    \includegraphics[width=\linewidth]{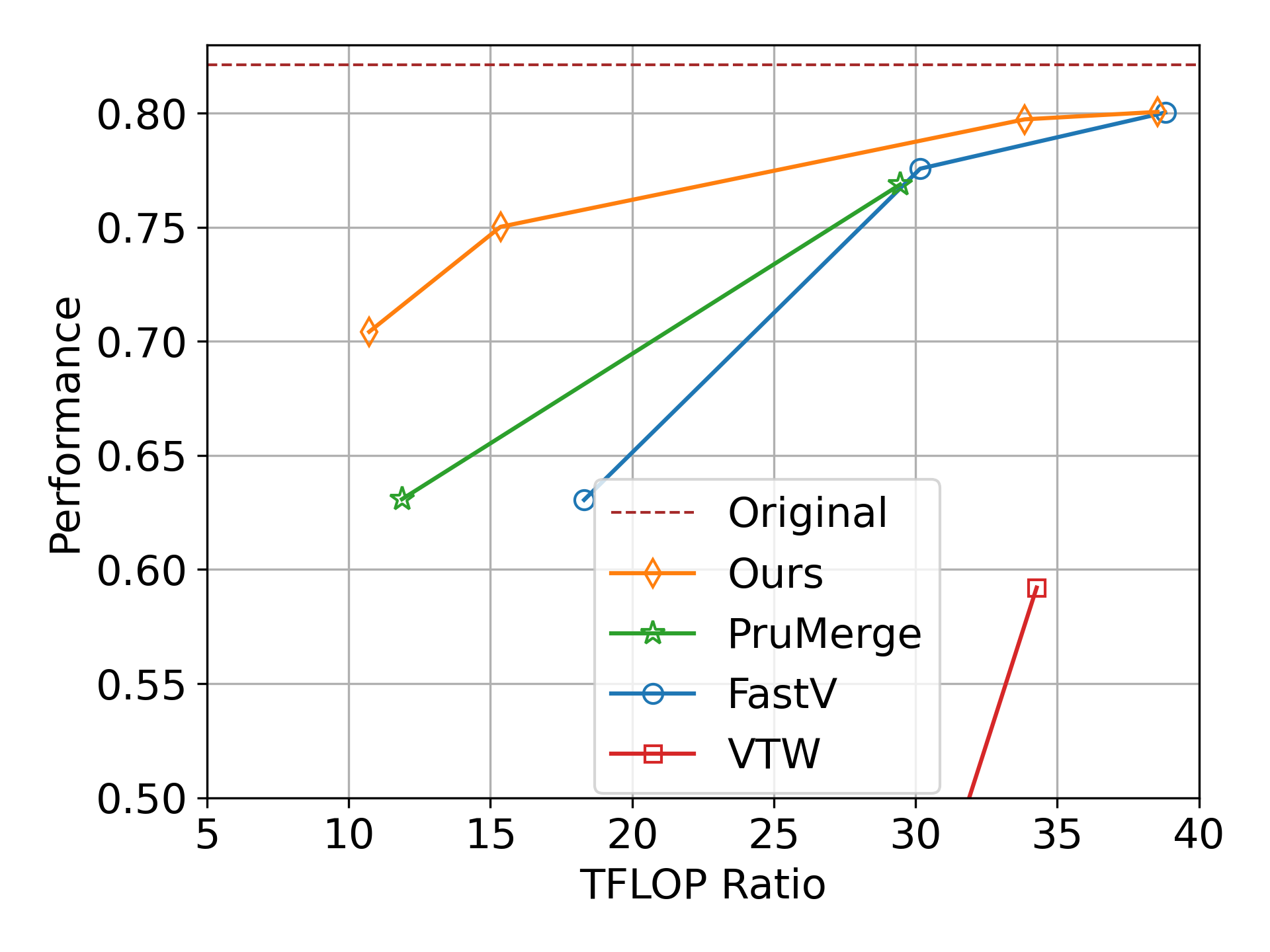}
    \caption{
    {Comparison of different visual token pruning methods across various pruning ratios for LLaVA 1.5-13B. The y-axis is the performance averaged on COCO (CIDEr), OKVQA (Acc), POPE (F1), and MMBench (Acc). The x-axis is the TFLOP ratio of the model after token pruning compared to the original model before pruning.} 
    }
    \label{fig:tradeoff_13b}
\end{figure}

\begin{figure*}[tbh!]
    \centering
    \begin{subfigure}[b]{0.49\textwidth}
        \includegraphics[width=\textwidth]{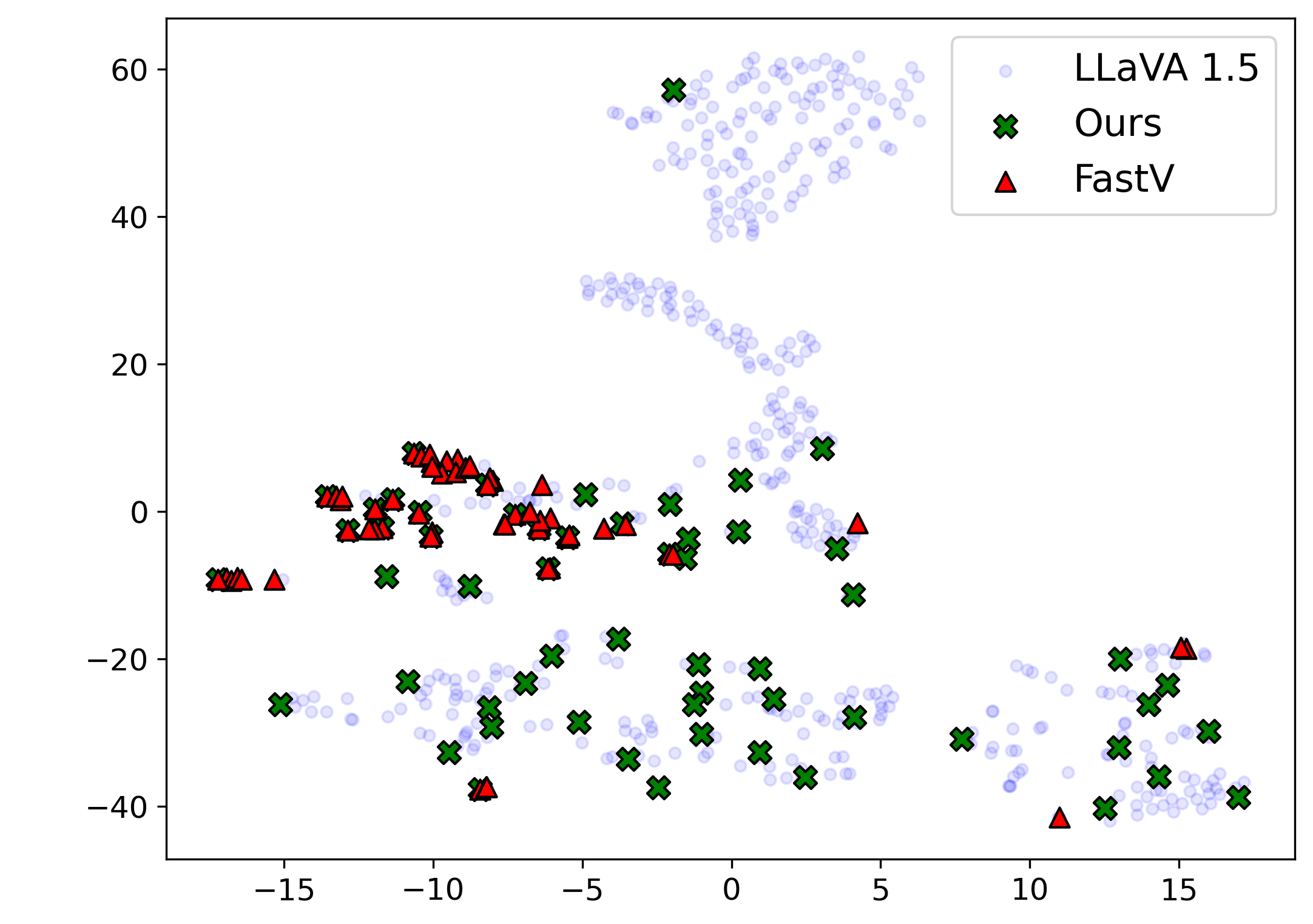}
        \caption{SeedBench example \#1}
        \label{fig:SeedBench-first}
    \end{subfigure}
    \hfill
    \begin{subfigure}[b]{0.49\textwidth}
        \includegraphics[width=\textwidth]{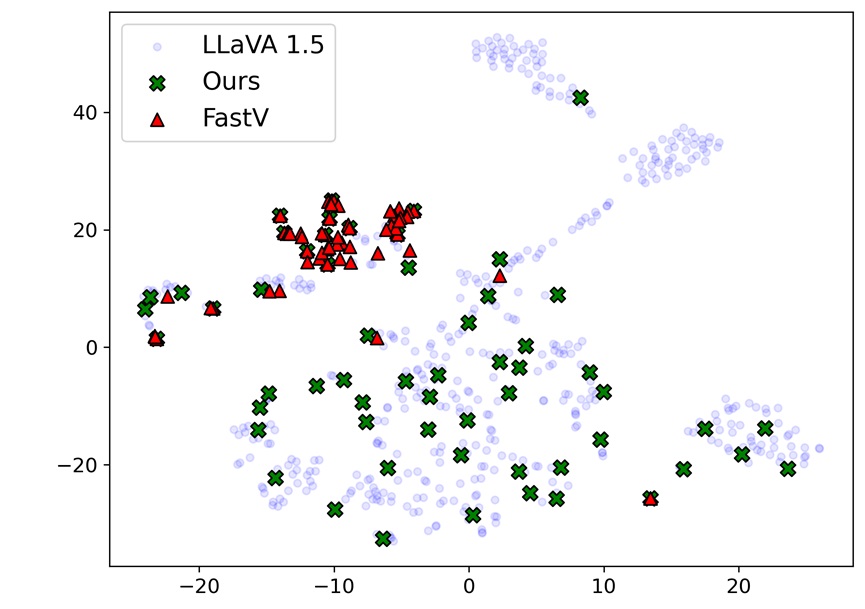}
        \caption{SeedBench example \#2}
        \label{fig:SeedBench-second}
    \end{subfigure}
    
    \vskip\baselineskip
    
    \begin{subfigure}[b]{0.49\textwidth}
        \includegraphics[width=\textwidth]{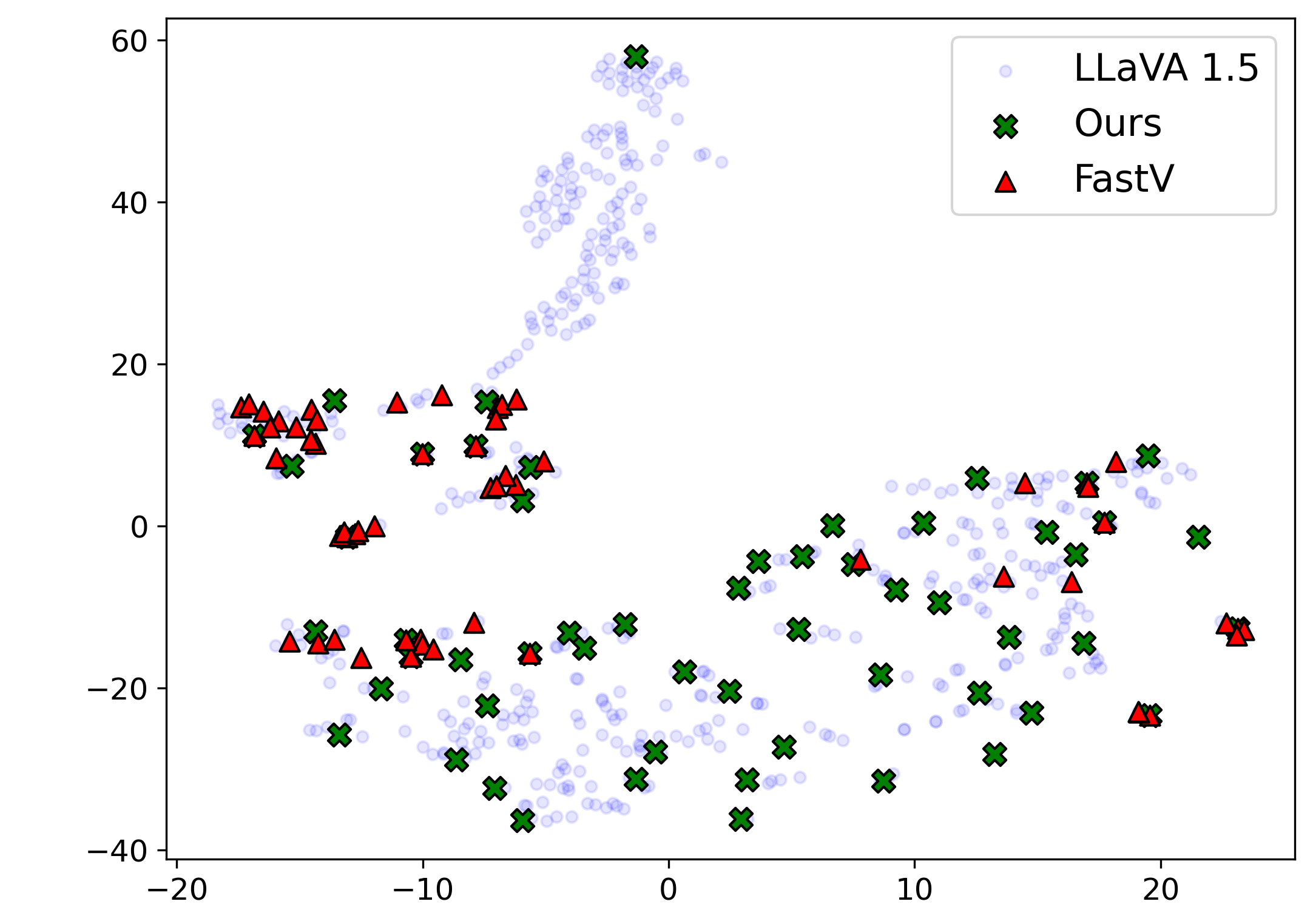}
        \caption{GQA example \#1}
        \label{fig:gqa-first}
    \end{subfigure}
    \hfill
    \begin{subfigure}[b]{0.49\textwidth}
        \includegraphics[width=\textwidth]{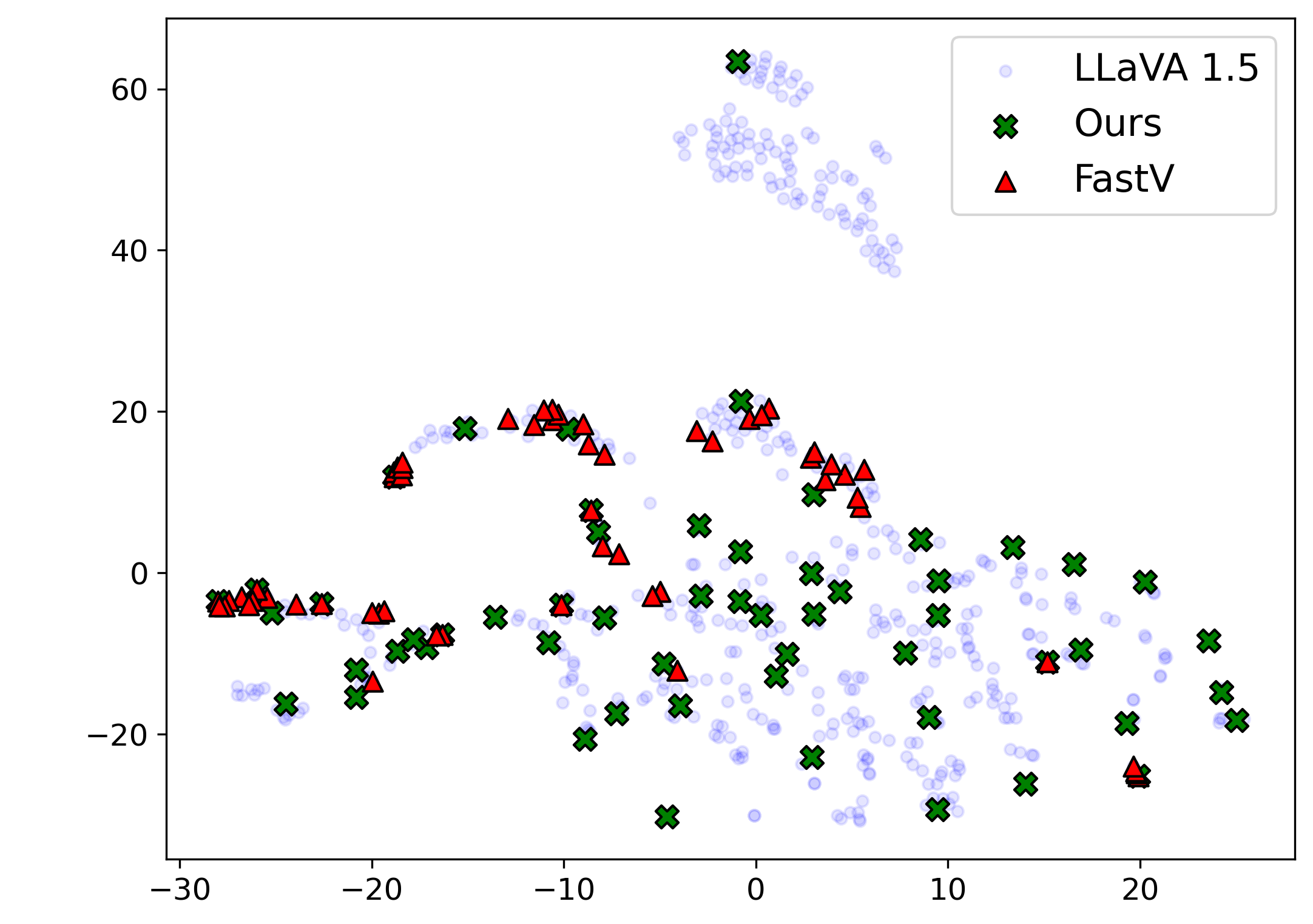}
        \caption{GQA example \#2}
        \label{fig:gqa_second}
    \end{subfigure}
    
    \vskip\baselineskip
    
    \begin{subfigure}[b]{0.49\textwidth}
        \includegraphics[width=\textwidth]{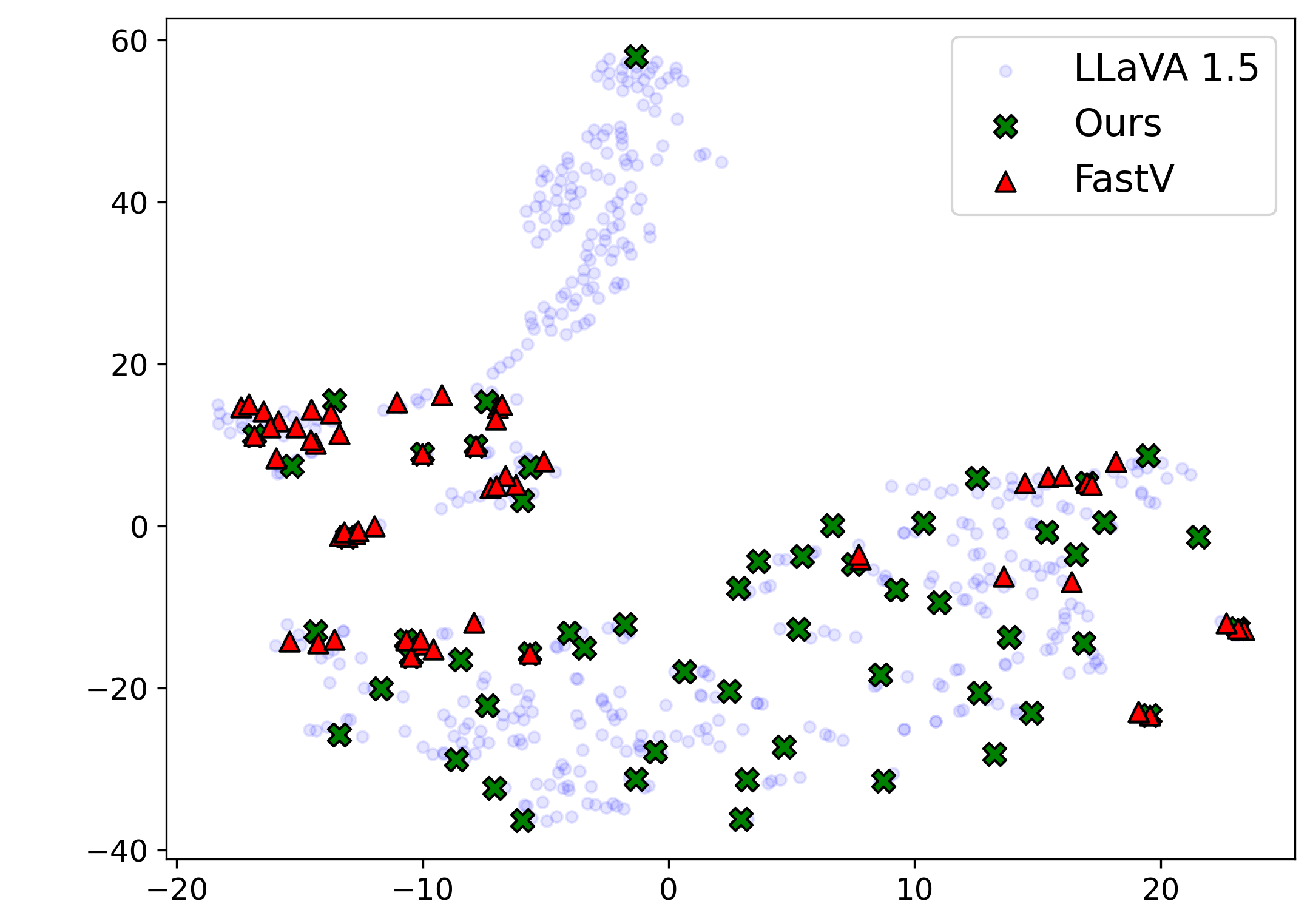}
        \caption{GQA example \#3}
        \label{fig:gqa_third}
    \end{subfigure}
    \hfill
    \begin{subfigure}[b]{0.49\textwidth}
        \includegraphics[width=\textwidth]{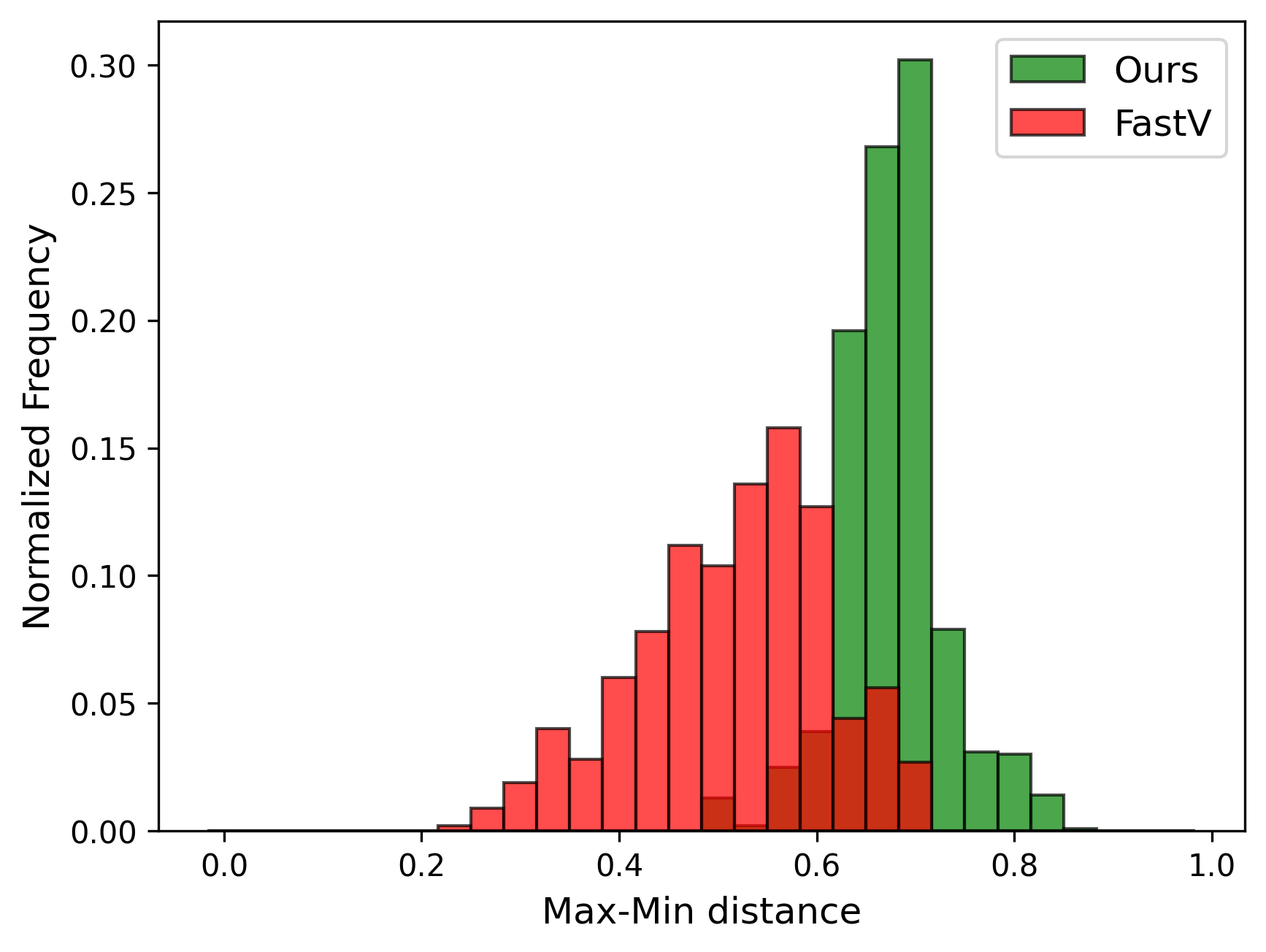}
        \caption{Histogram of Max-Min Distance for GQA dataset}
        \label{fig:gqa_histogram}
    \end{subfigure}
    
    \caption{(a)-(b) t-SNE visualization of visual tokens using SeedBench samples, (c)-(e) t-SNE visualization of visual tokens using GQA samples, (f) Histogram of the Max-Min distance between the selected tokens over the GQA dataset.}
    \label{fig:visualization_supp}
\end{figure*}

\section{Results with Additional Datasets}
In addition to the datasets tested in the main manuscript, we evaluate the proposed method and the baselines with LLaVA 1.5-7b model on more visual question answering datasets: TextVQA~\cite{textvqa}, VizWiz~\cite{vizwiz}, and VQAv2~\cite{vqa_v2}. The details corresponding to each dataset are included in Tab.~\ref{tbl:dataset_detail}. The same hyperparameters used for results in Tab.~1 of the main manuscript are applied to both our method and the baselines. The results for the proposed method and the baselines are summarized in Tab.~\ref{tbl:additional_dataset}. The TFLOPs are calculated for each dataset, and the average TFLOP and ratio are given in the TFLOP column of the table. VTW, FastV, and ours are the 3 training-free and calibration-free methods. As the results indicate, our method outperforms VTW and FastV on TextVQA, VizWiz, and VQAv2 datasets by $\approx$ 27\%, 7\%, and 29\%, respectively.

In the case of dynamic pruning scenario, we matched the pruning ratio with that of the PruMerge baseline \cite{PruMerge}. The comparison of our results with PruMerge reveals that our method achieves higher accuracy on VizWiz and VQAv2 datasets. Compared to FitPrune \cite{FitPrune}, which uses calibration datasets to optimize the procedure of token pruning, we achieve higher task performance on all the datasets. Finally, compared to the fine-tuning-based $\text{M}^{3}$ \cite{m3} method, our performance is worse on TextVQA, comparable on VQAv2, and better on VizWiz dataset. DivPrune achieves better results compared to the original model on VizWiz dataset. Visual token pruning has been shown to improve the original model's performance for some datasets~\cite{FastV}. Overall, the results shown in the table are inline with the results reported in the manuscript. This proves that DivPrune outperforms baselines on a diverse range of tasks and datasets.    

\subsection{Different TFLOPs for the 13b Model}
In the main manuscript, we showed the performance of baselines and our method across various TFLOP ratios for LLaVA 1.5-7b model. In this section, we present the results with LLaVA 1.5-13b model. The results are shown in Fig.~\ref{fig:tradeoff_13b} where the y-axis represents average performance on four datasets, namely, COCO (CIDEr), OKVQA (Acc), POPE (F1), and MMBench (Acc). For all datasets, the performance metric spans from 0 to 1, with the exception of the CIDEr metric, which can reach a peak value of 1.16 for the tested model. On the x-axis, we only show the high compression scenario (TFLOP ratio $\le$ 40\%). As shown in the figure, our method significantly outperforms all the baselines, particularly in high compression scenarios (TFLOP $\le$ 25\%). Furthermore, the gap between our approach and the baselines increases at extreme compression levels. For higher TFLOP ratios almost all methods converge toward the original performance. The pruning ratio and calibration samples for the FitPrune  are not provided for the 13b model, unlike the 7b model, hence it is excluded from the baselines.

\section{Qualitative Results}
\begin{figure*}[tbh!]
    \centering
    \begin{subfigure}{\linewidth}
        \centering
        \includegraphics[width=0.91\textwidth]{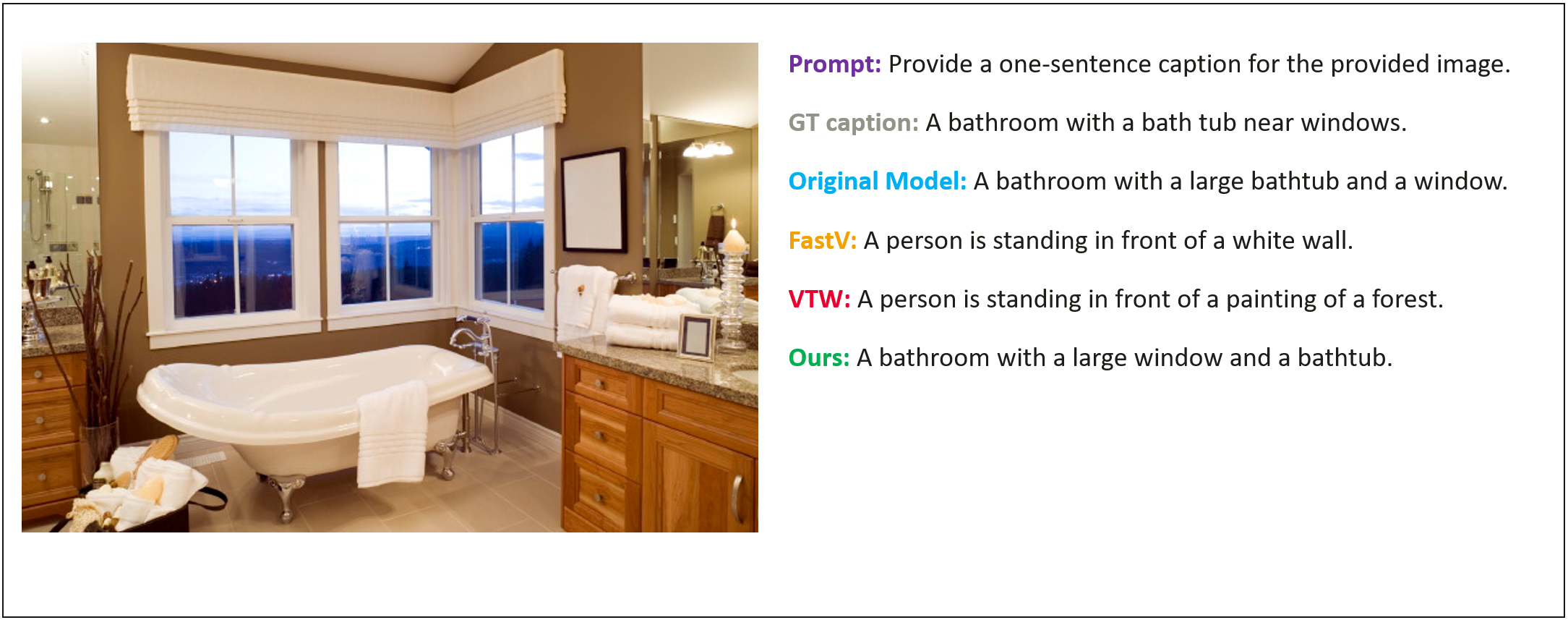}
        \caption{}
    \end{subfigure}
    
    \begin{subfigure}{\linewidth}
        \centering
        \includegraphics[width=0.91\textwidth]{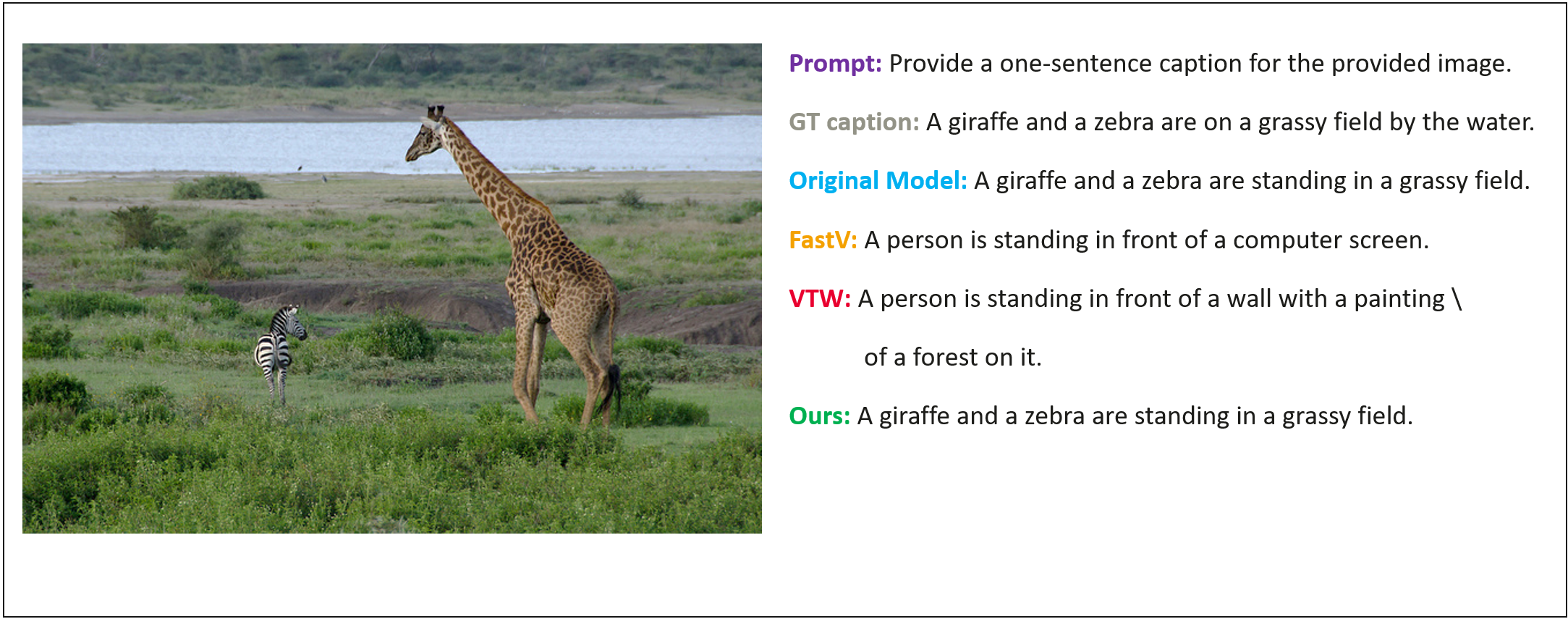}        
        \caption{}
    \end{subfigure}

    \begin{subfigure}{\linewidth}
        \centering
        \includegraphics[width=0.91\textwidth]{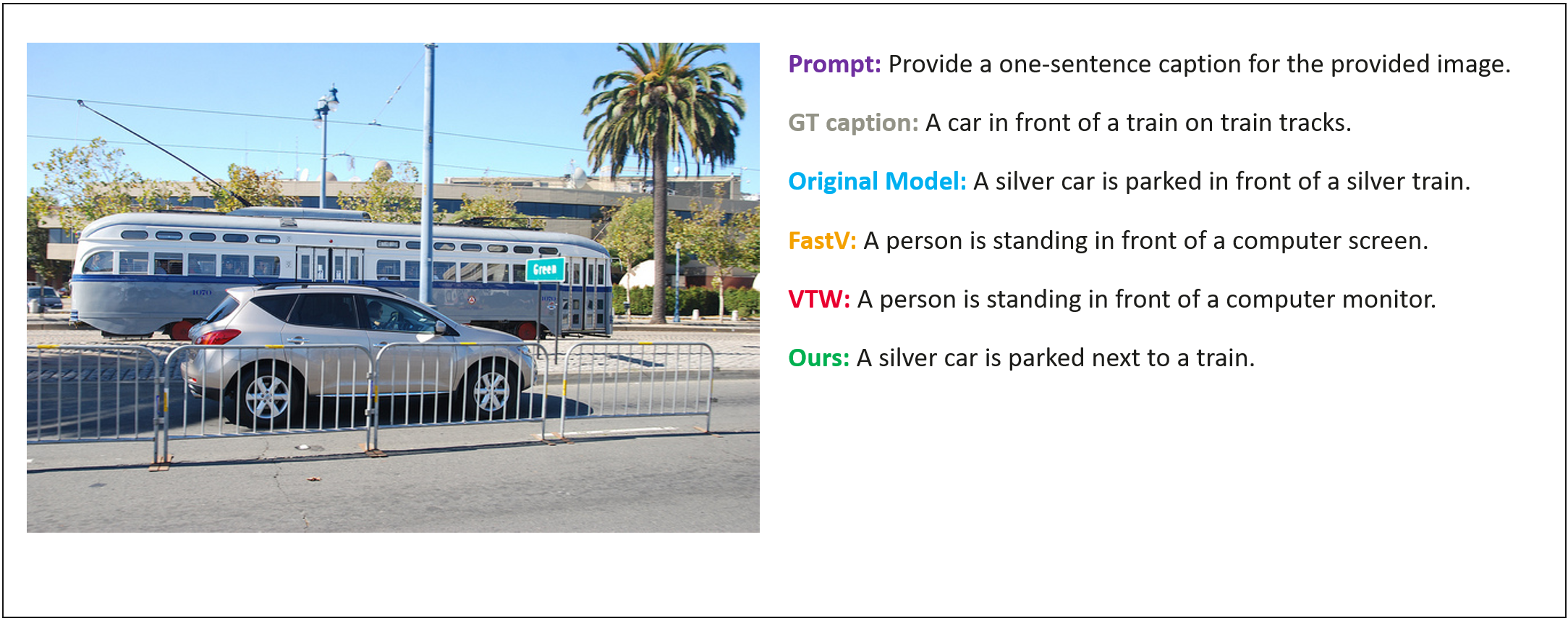}        
        \caption{}
    \end{subfigure}
    \caption{Visual examples for image captioning task, comparing the model outputs using the baselines and the proposed method. Colors in text are used for better readability.}
    \label{fig:qualitative_caption}
\end{figure*}

{In this section, we present some qualitative results comparing the proposed method with the relevant baselines.} Given the significant improvement of our method over the baselines on image captioning tasks, we provide 3 examples for image captioning using COCO \cite{coco} dataset in Fig.~\ref{fig:qualitative_caption}. For all the examples, the prompt, ground truth (GT) caption, and the LLaVA 1.5-7B model's output are given for reference. The model's output when our pruning method and baselines are applied is also shown for each example. We follow the experimental settings used to obtain the results in Tab.~1 of the main manuscript. The results show that using DivPrune (our method) enables the model to produce descriptions that closely align with the original model’s output, which is very similar to the ground truth, while only using 12\% TFLOP compared to the original model. In contrast, FastV and VTW generate irrelevant captions for the given images with the same TFLOP ratio. 

\begin{figure*}[tbh!]
    \centering
    \begin{subfigure}{\linewidth}
        \centering
        \includegraphics[width=0.91\textwidth]{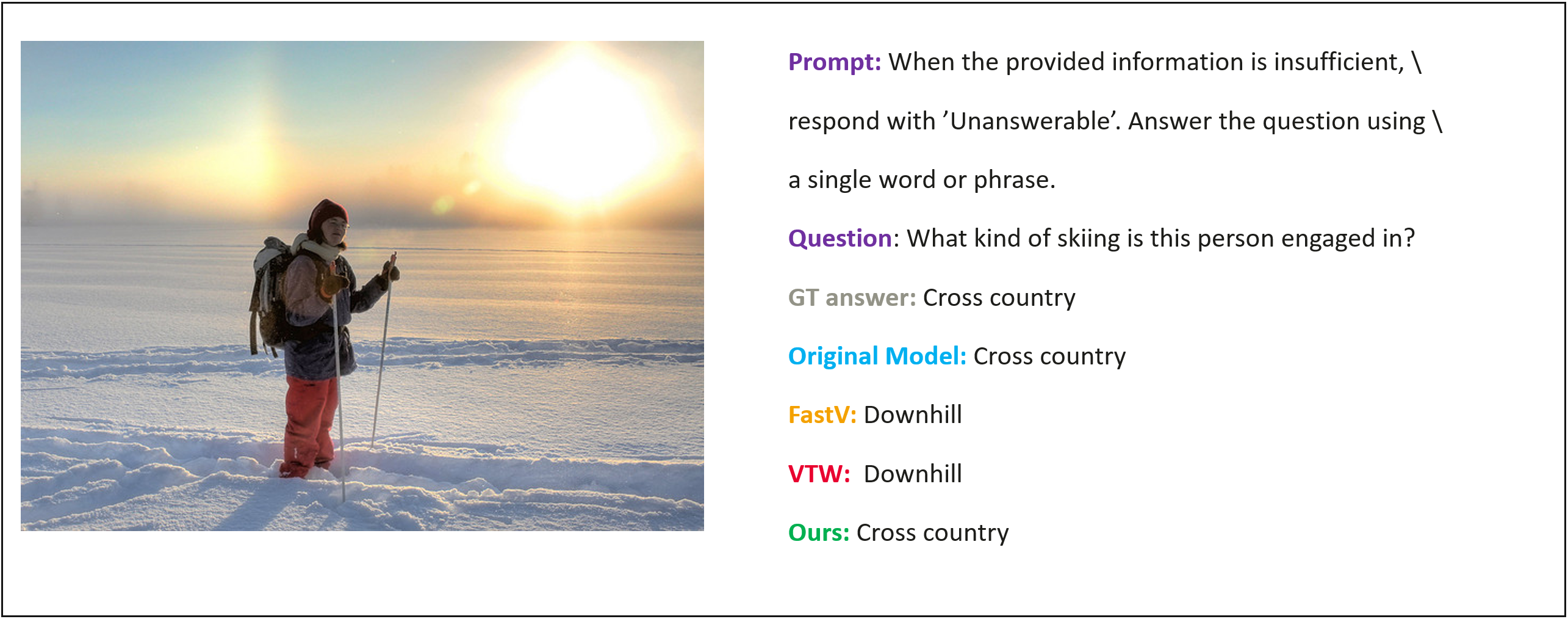}
        \caption{}
    \end{subfigure}
    
    \begin{subfigure}{\linewidth}
        \centering
        \includegraphics[width=0.91\textwidth]{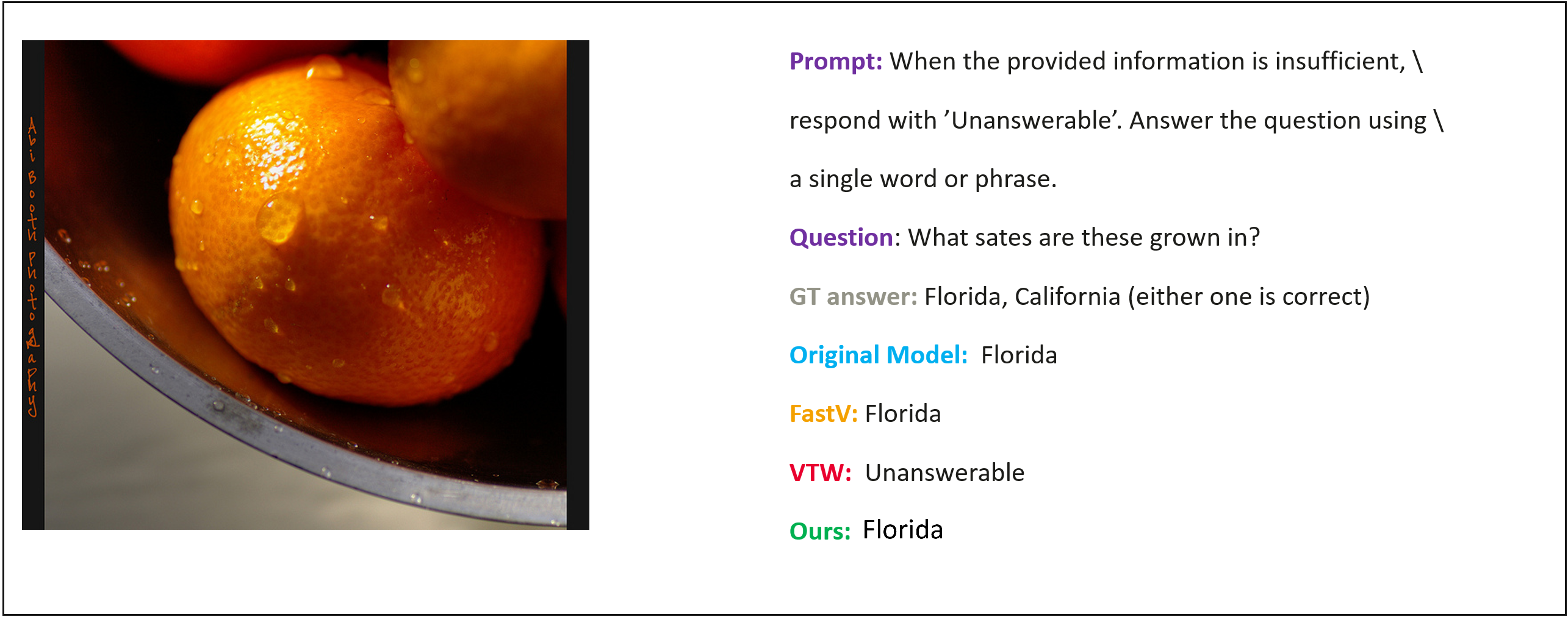}        
        \caption{}
    \end{subfigure}

    \begin{subfigure}{\linewidth}
        \centering
        \includegraphics[width=0.91\textwidth]{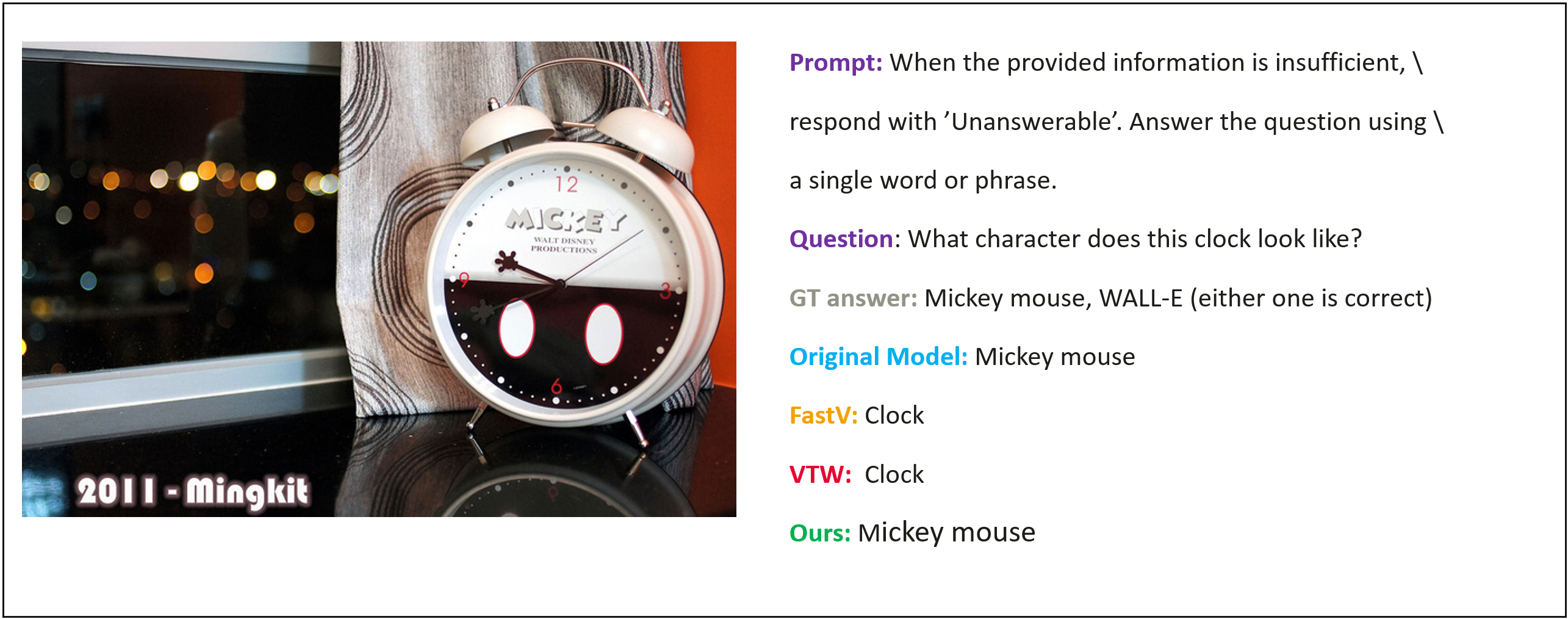}        
        \caption{}
    \end{subfigure}
    \caption{Visual examples for visual question answering task, comparing the model outputs using baselines and the proposed methods. Colors in text are used for better readability.}
    \label{fig:qualitative_VQA}
\end{figure*}

We also provide qualitative examples for a VQA task. Specifically, the output of LLaVA 1.5-7B model for sample images and questions from OKVQA~\cite{okvqa} dataset, along with the ground truth and the corresponding prompt are shown in Fig.~\ref{fig:qualitative_VQA}. As the figure illustrates, the output of the model when DivPrune is applied matches the ground truth. However, when FastV or VTW method are used, the model either generates incorrect responses or indicates that insufficient information is provided.

\section{Hyper-Parameters of Baselines}
In the main paper, TFLOP ratio and values are reported for DivPrune and the baselines. In this section, we provide the details on the hyperparameters specific to these methods. For DivPrune, the pruning ratio is set to 90.2\%. For FastV with 7B models $K=3$ and $R=0.001$, and with 13B models $K=3$, $R=0.023$ are used. For VTW, we use $K=4$ for LLaVA 1.5 models and $K=3$ for LLaVA 1.6 model. For $M^3$, $S$ is set to 56, and for FitPrune pruning ratio is set to 90\%. 

\clearpage
\newpage
\newpage
{
    \small
    \bibliographystyle{ieeenat_fullname}
    \bibliography{main}
}


\end{document}